\documentclass[10pt]{article}

\usepackage[margin=1in]{geometry}
\usepackage{times}
\usepackage{graphicx}
\usepackage{amsmath,amssymb}
\usepackage{booktabs}
\usepackage{multirow}
\usepackage{listings}
\usepackage{tabularx}
\usepackage{natbib}
\usepackage{url}
\usepackage{hyperref}
\usepackage{seqsplit}
\usepackage{caption}
\usepackage{float}
\usepackage{authblk}
\usepackage{longtable}

\newcommand{\code}[1]{\texttt{#1}}

\title{PolymerGPT: A Decoder-Based Generative Model for Inverse Polymer Design Conditioned on Properties and Scaffolds}
\author[1]{Charlie Pyle \thanks{Equal contribution}}
\author[2]{Adarsh Gadari $^*$}
\author[3]{C. Adrian Figg}
\author[4]{Zhenquan Jia}
\author[5]{Yaohang Li}
\author[5]{Chunjiang Zhu \thanks{Corresponding author: czhu@odu.edu}}

\affil[1]{Texas A\&M University}
\affil[2]{University of Pittsburgh}
\affil[3]{Department of Chemistry, Virginia Tech}
\affil[4]{Department of Biology, University of North Carolina Greensboro}
\affil[5]{Department of Computer Science, Old Dominion University}
\date{}

\begin{document}

\maketitle

\begin{abstract}

Polymer property prediction and inverse generative design targeting desired properties are important components of machine learning-assisted polymer design. Inverse design is particularly challenging when candidate materials must meet several property requirements.  In this paper, we present PolymerGPT, a decoder-only generative pre-trained transformer (GPT) model that generates polymer repeat units conditioned on property targets and, optionally, a specified scaffold. The model incorporates property information into the generative process through learned conditioning prefixes and accepts up to 37 property inputs. We evaluate unconditional, property-conditioned, and scaffold-conditioned generation, as well as the effects of model size, training-set size, and generation size. PolymerGPT achieves superior performance for conditional and unconditional generations while usually maintaining high validity, uniqueness, and novelty. In a five-property design demonstration, we specify targets for glass-transition temperature, bulk and chain electronic band gaps, electron affinity, and crystallinity, while setting the remaining 32 property inputs to their training-set means. Top generated candidates show close agreement with the five targets according to surrogate predictions. The conditioning framework can be extended to investigate additional property combinations.

\end{abstract}

\section*{Keywords}

Polymer Inverse Design; Conditional Generation; Polymer Property Condition; Scaffold Condition; Generative Pretrained Transformer

\section{Introduction}

Polymers are an integral part of our everyday lives, essential in building materials such as plastics, fibers, and rubbers. The sheer magnitude and diversity of the polymer chemical space offers opportunities to design polymers that match application demands, yet also pose the challenge of efficiently navigating this vast space. Traditional experimental approaches to novel polymer design have historically relied on a trial-and-error approach, requiring intensive time and resources while offering limited access to the vast polymer space. A new paradigm, driven by artificial intelligence (AI) and machine learning (ML), is accelerating polymer design through database construction, feature representation, development of ML-assisted property prediction models, generation and virtual screening of potential candidates, and experimental validation \cite{yue2025machine}.

Two key challenges remain within this framework.
First, unlike small-molecule drug discovery, where substantial libraries of candidate compounds exist, there has been limited availability of large datasets of polymer structures and properties. Second, while the development of ML models for polymer property prediction has received considerable attention, there have been limited studies on the generation of virtual polymer structures.  This may be due to the diversity of polymer structures, the broad range of their properties, and the limited availability of training samples, which add complexity to such generative models. Polymers consist of long-chain molecules composed of repeating units, whose specific arrangements and bonding patterns largely determine the macroscopic properties of the material. Carefully designing these repeating units results in materials with targeted functionalities and physical properties. 
Yue et al. \cite{BenchmarkInversePolymerDesign} 
evaluated unconditional generation (i.e., no targeted property) for deep generative models such as variational autoencoders. Using reinforcement learning, they then trained selected models for conditional generation on a desired high glass transition temperature ($T_g$). PolyTAO generates polymers from an initial template polymer with desired properties, similar to PolyG2G \cite{PolyG2G}; yet identifying such templates is challenging \cite{POLYTAO}. Recently, the chemical language model PolyT5 was built and adapted for polymer generation, but focusing on single‑property optimization in the generative process \cite{POLYT5}.

In the aforementioned works, typically only one property is optimized at a time, even though several physical characteristics are required to effectively predict macroscopic material behavior and utility. For example, designing dielectric polymers requires not only a high dielectric constant but also a wide band gap and high $T_g$ to ensure thermal and electrical stability, and adequate solubility for easy synthesizability. One approach is to generate a pool of candidate structures and then screen their predicted properties against the desired criteria. The effectiveness of this generate-and-screen approach depends in part on whether the generated pool contains candidates that satisfy the combined requirements. An alternative is to incorporate property targets into the generation process. Jointly designing polymers against many properties simultaneously—while maintaining high validity, uniqueness, and novelty—remains a key challenge.

\begin{figure*}[t]
    \centering
    \includegraphics[width=1\linewidth]{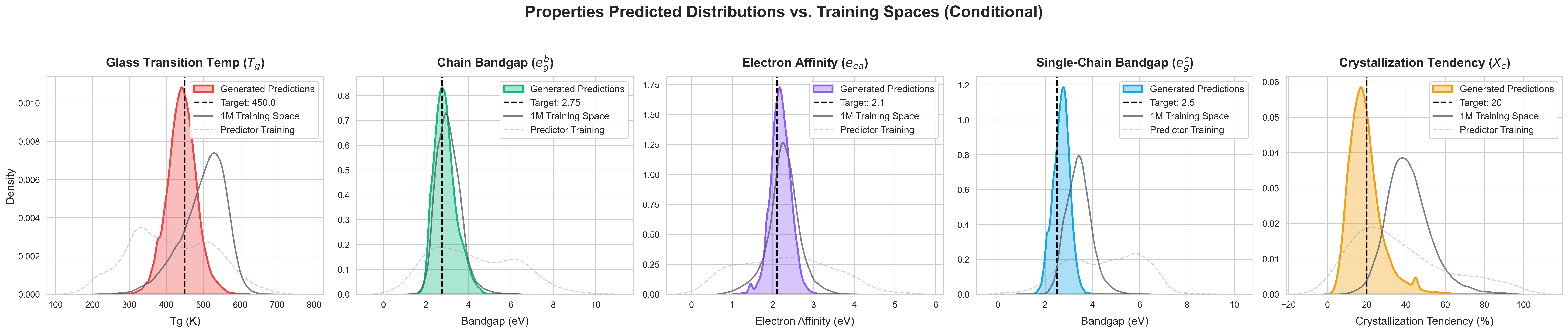}
    \caption{Distributions of TransPolymer-predicted values for the five properties targeted in the PolymerGPT design demonstration. The remaining 32 property inputs were set to their training-set means. Direct optimization of the five properties with respective predicted properties simultaneously centered at desired values.}
    \label{fig:cond_37props}
\end{figure*}

In this work, we present PolymerGPT, a decoder-only GPT model for generating polymer repeat units under property and scaffold conditions. PolymerGPT incorporates property information through learned conditioning prefixes and accepts up to 37 polymer property inputs. We systematically evaluate its generation capabilities for unconditional, property-conditioned, and scaffold-conditioned generations and examine the effects of training-set size, model size, and generation size. Our results demonstrate that PolymerGPT achieves decent generative performance for conditional and unconditional generations across diverse tested setups. In particular, we specify targets for five properties and set the remaining 32 inputs to their training-set means. Selected generated candidates have TransPolymer-predicted properties close to the five specified targets ({\bf Figures~\ref{fig:cond_37props} and \ref{fig:top10fig}}). Our method also holds promises for studying additional property combinations.

\section{Related Work}
Recent work on data‑driven polymer design spans property prediction models, deep generative models, and conditional chemical language models. On the {\em property prediction} PolyNC introduces a joint natural‑language and chemical‑language model that combines free‑text property descriptions with pSMILES‑like polymer sequences to predict a unified set of polymer properties, leveraging natural‑language prompts to capture semantic relationships between properties \cite{PolyNC}. PolyBERT, in turn, treats pSMILES as a chemical language and uses a BERT‑style architecture trained on tens of millions of polymers to learn dense fingerprints that enable property prediction and screening over large polymer spaces \cite{polyBERT}. TransPolymer introduces a RoBERTa‑style Transformer pretrained via masked language modeling on 5 million polymer sequences and fine‑tuned across ten downstream property datasets; as a discriminative encoder, it maps polymer sequences to scalar property values \cite{TransPolymer}. On the {\em generative} side, a systematic benchmark compares six deep generative architectures such as Variational Autoencoder, evaluating their ability to generate chemically valid and diverse polymers and, via reinforcement learning, to steer candidates toward improved task‑specific properties \cite{BenchmarkInversePolymerDesign}. MolGPT introduces a Transformer‑decoder for de novo molecular generation using SMILES strings and demonstrates the viability of decoder‑only architectures for conditional generation, focusing on drug‑like small molecules  \cite{MolGPT}. PolyTAO proposes a Transformer‑based pretrained model, generating polymers from user‑specified property targets and supporting both semi‑template and template‑free generation over a curated structure–property dataset, with extensions to additional properties via task‑specific fine‑tuning \cite{POLYTAO}. PolyG2G implements a graph-to-graph translation framework to recover established structure–property relationships, achieving high performance results \cite{PolyG2G}. In \cite{batra2020polymers}, a syntax-directed variational autoencoder (SD-VAE) is used to design polymers for extreme thermal and mechanical conditions, using a grammar-constrained latent space to help ensure generated structures are chemically valid. Within \cite{liu2023high}, an invertible graph generative model (IGGM) is used to design high-temperature polymer dielectrics, applying reversible graph transformations to explore the structure–property space. Mole. Chef, proposed in \cite{kim2023open}, is a fragment-based generative framework restricted to synthetically accessible monomer combinations. Graph DiT is a graph diffusion Transformer for multi-conditional polymer and small-molecule generation but it focused on targeting polymers with high O2 and low N2 permeability \cite{liu2024graph}. polyBART introduces a BART-based chemical language model built on Pseudo-polymer SELFIES (PSELFIES), unifying property prediction and single-property generative design within one encoder–decoder architecture, with a designed polymer experimentally validated for high thermal degradation temperature \cite{savit2025polybart}. The inverse design and property prediction tasks were both merged within PolyT5, which presents an encoder–decoder foundation chemical language model built on the T5 architecture \cite{POLYT5}.

% ─────────────────────────────────────────────────────────────────────────────
\section{PolymerGPT for Multi-Property Inverse Polymer Generation}

\begin{figure*}[t]
    \centering
    \includegraphics[width=\linewidth]{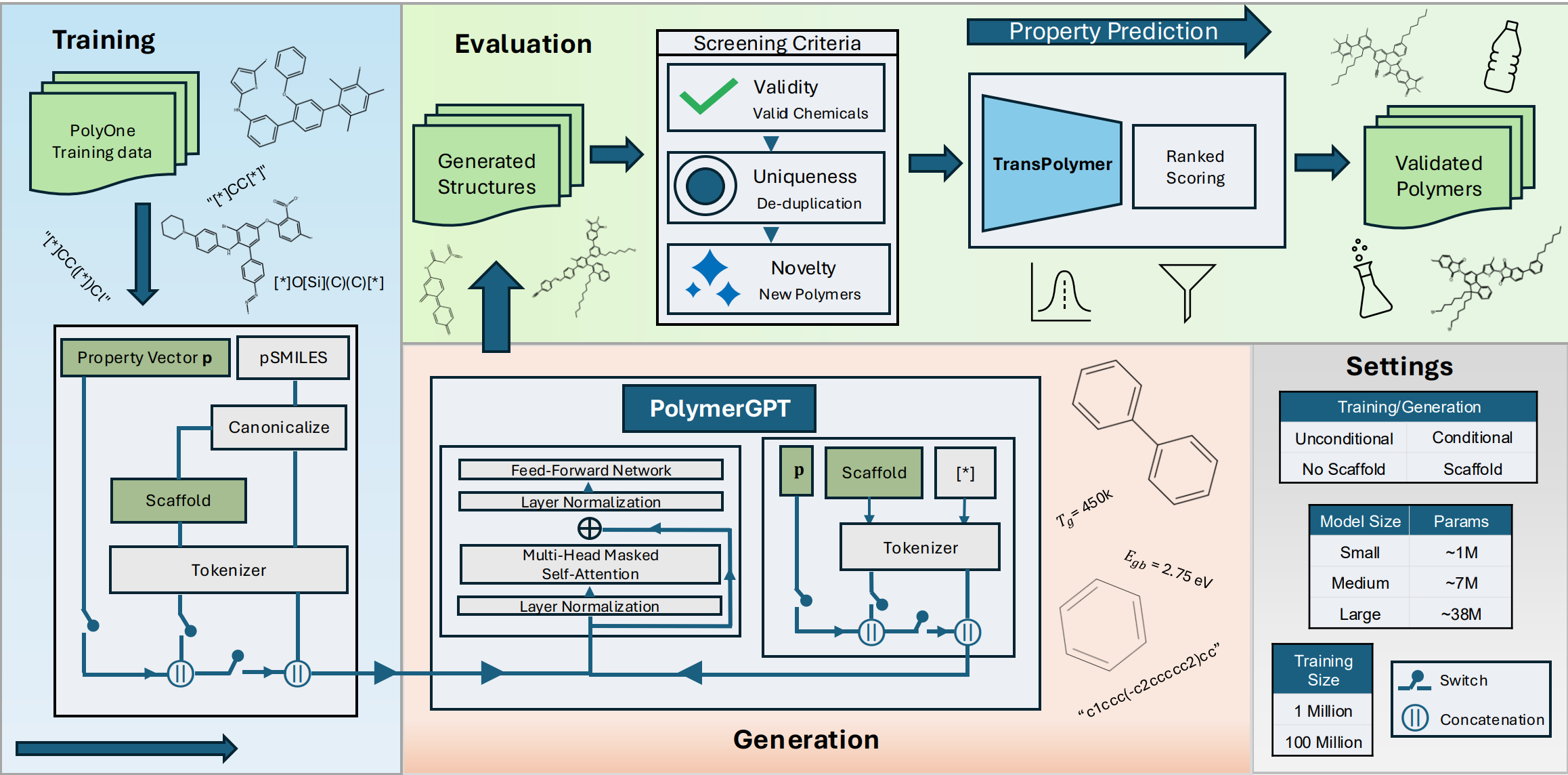}
    \caption{Overview of the PolymerGPT framework. During training, polymer repeat units from PolyOne are tokenized, with property vectors \textbf{p} and scaffold information optionally concatenated as conditioning prefixes to the pSMILES sequence. During generation, PolymerGPT autoregressively samples new pSMILES via masked self-attention, conditioned on property, scaffold, and 
    \texttt{[*]} wildcard tokens. Generated structures are filtered by validity, uniqueness, and novelty, before properties are predicted with the TransPolymer model (with optional ranked filtering). Model size, training size, and conditioning settings are configurable, as shown in the Settings panel.}
    \label{fig:architecture}
\end{figure*}

We begin by describing our generative modeling approach, along with the polymer data representation and tokenization scheme, followed by the details of our model architecture.

\subsection{Generative Modeling}
Since polymers are long-chain molecules that are difficult to represent in full, we use repeat units, which are a representative method for capturing essential structural information \cite{yue2025machine}. Building on this, we perform generation of polymer structures conditioned on multiple target physical properties. Given a desired property vector $\mathbf{p} \in \mathbb{R}^K$, the goal is to sample polymer SMILES sequences $\mathbf{x} = (x_1, \ldots, x_N)$ from a learned distribution $P_\Theta(\mathbf{x} \mid \mathbf{p})$ such that generated structures exhibit properties near the target $\mathbf{p}$. We model $P_\Theta$ with an autoregressive Transformer decoder, factorizing the sequence likelihood as $\prod_{i=1}^{N} P_\Theta(x_i \mid x_{<i}, \mathbf{p})$. We inject $\mathbf{p}$ as a conditioning prefix, prepended directly to the token sequence, so that the property signal participates in the same masked self-attention computation used for next-token prediction.

\subsection{Data Representation, Tokenization, and Properties}
\label{sec:data}

{\noindent \bf Data} We use the largest publicly available polymer dataset, PolyOne, which contains 100 million polymer structures, along with a subset of 1 million structures for model training. PolyOne consists of hypothetical polymers generated by enumerating combinations of chemical fragments extracted from more than 13,000 synthesized polymers \cite{polyone}. Each polymer structure is associated with 37 widely used polymer properties, where the property labels are predictions produced by PolyBERT-based property models \cite{polyBERT}. The 37 properties span multiple domains. Electronic properties include the chain and bulk electronic band gaps ($e_g^c$, $e_g^b$), electronic injection barrier ($e_{\mathrm{ib}}$), cohesive energy density ($\mathrm{ced}$), ionization energy  ($e_i$), electron affinity ($e_{\mathrm{ea}}$). Optical and dielectric properties include refractive index at DFT and experiment ($n_c$, $n_e$), dielectric constants at nine frequencies ($\varepsilon_{\mathrm{se},*}$; $10^{1.78--15.0}$ Hz), and chain dielectric constant ($\varepsilon_c$). Mechanical properties include tensile strength at break ($\mathrm{tsb}$) and at yield ($\mathrm{tsy}$), elongation at break ($\varepsilon_b$), and Young's modulus ($\mathrm{ym}$). Permeability properties comprise gas permeability for six gases ($\mathrm{P}*$: CH$_4$, CO$_2$, H$_2$, O$_2$, N$_2$, He). Thermodynamic and physical properties include atomisation energy ($E_{\mathrm{at}}$), density ($\rho$), limiting oxygen index ($\mathrm{loi}$),
crystallization tendency from DFT and experiments ($X_c$, $X_e$), and heat capacity ($c_p$). Thermal properties are decomposition temperature ($T_d$), glass-transition temperature ($T_g$), and melting temperature ($T_m$).
{\bf Table~\ref{tab:property_provenance}} provides the properties, symbols, and units in PolymerGPT and provenance of the underlying data used to train the PolyBERT property predictors. The frequency-dependent dielectric constant is one of the 29 property types in PolyOne; treating its nine frequency-specific values as separate inputs yields 37 property inputs. For the five properties in our main design demonstration, the glass-transition temperature predictor was trained on experimental data, whereas the predictors for the bulk and chain band gaps, electron affinity, and crystallization tendency were trained on DFT-derived data. Although predicted property values are not always accurate, our models learn robust latent structure–property relationships and generate plausible polymer structures.

\begin{longtable}{cllllrr}
\caption{Polymer properties used by PolymerGPT and provenance of the
underlying data used to train the PolyBERT property predictors.}
\small
\label{tab:property_provenance} \\
\toprule
ID & Symbol & Property & Unit & Source & Range & HP / CP \\
\midrule
\endfirsthead

\multicolumn{7}{l}{\textit{Table \thetable\ continued}} \\
\toprule
ID & Symbol & Property & Unit & Source & Range & HP / CP \\
\midrule
\endhead

\midrule
\multicolumn{7}{r}{\textit{Continued on next page}} \\
\endfoot

\bottomrule
\multicolumn{7}{p{0.95\linewidth}}{\footnotesize
\textit{Note:} HP and CP denote homopolymer and copolymer data points,
respectively. Sources, ranges, and counts are from PolyBERT
Table~1~\cite{polyBERT} and describe the datasets used to train its
property predictors. PolyOne property labels are model predictions.
A dash indicates that no CP count was reported. For dielectric
constants, PolyBERT reports an aggregate range and count for the
frequency-dependent property.} \\
\endlastfoot

\multicolumn{7}{l}{\textit{Electronic}} \\
1 & $e_g^c$ & Band gap (chain) & eV & DFT
  & 0.02--10 & 4,224 / --- \\
2 & $e_g^b$ & Band gap (bulk) & eV & DFT
  & 0.4--10 & 597 / --- \\
3 & $e_{ib}$ & Electronic injection barrier & eV & DFT
  & 2--7 & 2,610 / --- \\
4 & $\mathrm{ced}$ & Cohesive energy density & cal/cm$^3$ & Exp.
  & 20--300 & 294 / --- \\
5 & $e_i$ & Ionization energy & eV & DFT
  & 4--10 & 370 / --- \\
6 & $e_{ea}$ & Electron affinity & eV & DFT
  & 0.4--5 & 368 / --- \\

\addlinespace
\multicolumn{7}{l}{\textit{Optical and dielectric}} \\
7 & $n_c$ & Refractive index (DFT) & --- & DFT
  & 1--3 & 382 / --- \\
8 & $n_e$ & Refractive index (exp.) & --- & Exp.
  & 1--2 & 516 / --- \\
9--17 & $\varepsilon_{se,*}$ & Dielectric constant at freq. $f$
  & --- & Exp. & 2--10 & 1,187 / --- \\
18 & $\varepsilon_c$ & Dielectric constant, chain & --- & DFT
  & 3--9 & 382 / --- \\

\addlinespace
\multicolumn{7}{l}{\textit{Mechanical}} \\
19 & $\mathrm{tsb}$ & Tensile strength at break & MPa & Exp.
  & 0.005--200 & 663 / 318 \\
20 & $\mathrm{tsy}$ & Tensile strength at yield & MPa & Exp.
  & 0.00003--100 & 216 / 78 \\
21 & $\varepsilon_b$ & Elongation at break & \% & Exp.
  & 0.3--1,000 & 868 / 260 \\
22 & $\mathrm{ym}$ & Young's modulus & MPa & Exp.
  & 0.02--4,000 & 592 / 322 \\

\addlinespace
\multicolumn{7}{l}{\textit{Permeability}} \\
23 & $\mathrm{P}_{\mathrm{CH_4}}$ & CH$_4$ permeability
  & Barrer & Exp. & 0.0004--2,000 & 331 / 47 \\
24 & $\mathrm{P}_{\mathrm{CO_2}}$ & CO$_2$ permeability
  & Barrer & Exp. & 0.000001--5,000 & 286 / 119 \\
25 & $\mathrm{P}_{\mathrm{H_2}}$ & H$_2$ permeability
  & Barrer & Exp. & 0.02--5,000 & 240 / 46 \\
26 & $\mathrm{P}_{\mathrm{O_2}}$ & O$_2$ permeability
  & Barrer & Exp. & 0.000005--1,000 & 390 / 210 \\
27 & $\mathrm{P}_{\mathrm{N_2}}$ & N$_2$ permeability
  & Barrer & Exp. & 0.00003--500 & 384 / 99 \\
28 & $\mathrm{P}_{\mathrm{He}}$ & He permeability
  & Barrer & Exp. & 0.05--2,000 & 239 / 58 \\

\addlinespace
\multicolumn{7}{l}{\textit{Thermodynamic and physical}} \\
29 & $E_{at}$ & Atomization energy & eV/atom & DFT
  & $-7$ to $-5$ & 390 / --- \\
30 & $\rho$ & Density & g/cm$^3$ & Exp.
  & 0.8--2 & 910 / --- \\
31 & $\mathrm{loi}$ & Limiting oxygen index & \% & Exp.
  & 10--70 & 101 / --- \\
32 & $X_c$ & Crystallization tendency (DFT) & \% & DFT
  & 0.1--100 & 432 / --- \\
33 & $X_e$ & Crystallization tendency (exp.) & \% & Exp.
  & 1--100 & 111 / --- \\
34 & $c_p$ & Heat capacity & J/(g\,K) & Exp.
  & 0.8--2 & 79 / --- \\

\addlinespace
\multicolumn{7}{l}{\textit{Thermal}} \\
35 & $T_d$ & Degradation temperature & K & Exp.
  & 300--1,000 & 3,584 / 1,064 \\
36 & $T_g$ & Glass-transition temperature & K & Exp.
  & 80--900 & 5,183 / 3,312 \\
37 & $T_m$ & Melting temperature & K & Exp.
  & 200--900 & 2,132 / 1,523 \\

\end{longtable}

{\noindent \bf Tokenization }Polymer SMILES (pSMILES) strings use [*] wildcards to denote repeating unit attachment points. Each pSMILES string is canonicalized and tokenized with the SentencePiece tokenizer offered by PolyBERT \cite{kudo2018sentencepiece}. The tokenizer has a vocabulary of 265 tokens, including
common pSMILES characters such as the uppercase and lowercase forms of 118 elements of the periodic table of elements, numbers ranging from 0 to 9, and special characters like [*], (, ), =, among others. Sequences are padded to the maximum token length $l_{\max} = 397$, which is sufficiently large to encode the pSMILES strings in the datasets.

\subsection{Generative Models}
\label{sec:architecture}

PolymerGPT supports unconditional, conditional, and scaffolding generations. The overall training and generation workflow is illustrated in {\bf Figure~\ref{fig:architecture}}. For unconditional training, polymer pSMILES are tokenized using the SentencePiece tokenizer, and the model is trained on a next-token prediction objective. For conditional training, polymer property names and values are provided as conditioning inputs alongside the pSMILES tokens. During generation, the model receives a start token, which we use [*], and an optional set of property conditions, and then autoregressively predicts the next token to produce a polymer structure consistent with the specified conditions.

Our model follows the standard decoder-only Transformer architecture used in GPT models, which has demonstrated outstanding performance in various generative tasks. We are also inspired by MolGPT, which we adapt to the polymer domain by operating natively on pSMILES and enabling simultaneous property conditioning. Specifically, PolymerGPT comprises $L$ stacked decoder blocks, each of which is composed of a masked self-attention layer ($\mathrm{Attention}$) and a fully connected neural network ($\mathrm{FNN}$). Each self-attention layer returns a vector of size $d$ that is taken as input by the fully connected network. The $\mathrm{FNN}$ is a two-layer MLP with hidden dimension $4d$ and GELU activation. Normalization is applied before both the self-attention and $\mathrm{FNN}$ layers, which stabilizes training at larger depth. A residual connection is added for each layer's output and a dropout rate of $p$=0.1 is applied. The output of the last layer of $\mathrm{FNN}$, after the residual sum, is used as input for the next decoder block. The computation of each block is as follows:
{\fontsize{9}{\baselineskip}\selectfont
\begin{align}
  \mathbf{x} &\leftarrow \mathbf{x} + \mathrm{Attention}\!\left(\mathrm{LayerNorm}(\mathbf{x})\right) \\
  \mathbf{x} &\leftarrow \mathbf{x} + \mathrm{FNN}\!\left(\mathrm{LayerNorm}(\mathbf{x})\right)
\end{align}
}

{\noindent \bf Conditioning Prefix}
PolymerGPT incorporates multiple properties directly into the generative process through a learned conditioning prefix, allowing flexible multi‑property targeting without templates or post‑hoc RL reweighting. 
In addition to pSMILES tokens, there are two extra sets of tokens: \emph{position tokens} that indicate the position of each token in a pSMILES string, and \emph{type tokens}, which help the model differentiate condition tokens (type $0$) from polymer pSMILES tokens (type $1$). All tokens are transformed to an $d$-dimensional vector using a separate embedding layer. In particular, multi-property conditions are mapped to an $d$-dimensional vector through a fully connected linear layer. These pSMILES token embeddings, position embeddings, and type token embeddings are then added, and the resulting vector is then passed as input to the model. We study small, medium, and large model sizes of $\sim$1M, 7M, and 38M parameters with detailed setups in Table~\ref{tab:variants}

Our conditioning method accommodates the 37 properties available in the training dataset. In the five-property demonstration, we assign the desired values to five inputs and use training-set mean values for the remaining 32. We note that mean values in the training data are provided as the {\em default} property values, and that users can provide desired values for conditioned properties to override the default values. This allows users to input desired values for only those properties they aim to optimize, without having to specify additional values.

\smallskip
{\noindent \bf Scaffold Condition}
We introduce another type of condition, {\em scaffold condition}, that allows users to specify a desired scaffold for predicted structures. The pSMILES string of a scaffold will be canonicalized and tokenized, and the tokens will be transformed to $d$-dimensional vectors through the same embedding layer used for polymer pSMILES tokens. The scaffold embedding, after summing with type token embeddings, will be prepended as the condition. PolymerGPT supports a combined use of scaffold and property conditions.

\section{Experimental Evaluations and Discussions}
We first present experimental methods and then discuss experimental results in unconditional, conditional, and scaffolding generations. We then report additional experimental studies conducted to evaluate the impacts of model size, training size, and generation size. 
\subsection{Experimental Methods}
{\noindent \bf Performance Measures} For evaluation, we compare model performance using multiple metrics. pSMILES validity (VAL) is defined as the fraction of generated strings that can be parsed and canonically represented as polymer repeat units, calculated using RDKit. Uniqueness (UNQ) is computed after canonicalization by checking whether a generated polymer is distinct from all other generated polymers. Novelty (NOV) measures the fraction of generated polymers whose canonical pSMILES representations are not present in the training set, indicating genuinely new polymers. These filters are applied stepwise: starting from all generated polymers, we take the VAL subset, then the UNQ subset (valid and non‑duplicate), and finally the NOV subset (valid, non‑duplicate, and not in the training data). Additionally, during scaffold generation, we calculate a scaffold match metric as a fraction of valid polymers. This uses RDKit's Bemis-Murcko core scaffold framework, stripping non-ring side chains and linkers to isolate the exact ring skeleton. Unlike raw string matching or substructure matching, which can mis-handle side chains and extra rings, Murcko scaffolds strictly enforce the core topology.

{\noindent \bf Property Predictors}
To evaluate the accuracy of our generated structures, we employ the TransPolymer model as our primary property predictor, which is one of the strongest and publicly available predictors \cite{TransPolymer}. For matching properties already included in TransPolymer, we fine‑tune the model using the datasets and parameters provided. For other properties, we use additional fine‑tuning datasets and perform hyperparameter optimization using Bayesian optimization with Hyperband, implemented via the RayTune library. Due to the computational cost of this tuning process, we restrict our study to a subset of 5 property predictors. Appendix~\ref{sec:appendix_transpolymer} specifies the fine‑tuning data sources, train/test sizes, and hyperparameters searched for and selected. In {\bf Table~\ref{tab:property_performance}}, we summarize the training and test statistics for our property prediction model. As shown in Table~\ref{tab:property_performance}, the $T_g$ predictor achieves a test $R^2$ of 0.904 and a test RMSE of 35.07 K, indicating good predictive performance on its held-out test set. The other predictors achieve test $R^2$ values above 0.90, except for crystallinity ($X_c$), for which the test $R^2$ is 0.44 and the test RMSE is 17.51. All our test $R^2$ are similar to those reported in the TransPolymer paper. These results provide context for interpreting the predicted properties of generated polymers.

\begin{table}[!htbp]
\centering
\fontsize{9}{12}\selectfont
\setlength{\tabcolsep}{4pt}
\renewcommand{\arraystretch}{1.05}
\begin{tabular}{l r r r r}
\hline
Symbol (ID) & Train RMSE & Test RMSE & Train $R^2$ & Test $R^2$ \\
\hline
1: $e_{g}^{c}$       & 0.1562  & 0.4371  & 0.9897 & 0.9217 \\
2: $e_{g}^{b}$       & 0.1741  & 0.5286  & 0.9906 & 0.9254 \\
6: $e_{\mathrm{ea}}$ & 0.1247  & 0.3305  & 0.9850 & 0.9015 \\
32: $X_{c}$          & 9.6564  & 17.5126 & 0.7865 & 0.4445 \\
36: $T_{g}$          & 11.7552 & 35.0666 & 0.9886 & 0.9035 \\
\hline
\end{tabular}
\caption{Property prediction model performance.}
\label{tab:property_performance}
\end{table}

In the following, we present the performance of PolymerGPT models trained unconditionally, conditionally on a single property, on 5 properties, and with scaffolding. Unless otherwise stated, all results are reported using the medium-sized model and the dataset of 1 million polymers. Generation sizes target $\sim$50,000 polymers for each individual sweep.

\subsection{Unconditional Generation}

\begin{figure*}[t]
    \centering
    \includegraphics[width=\textwidth]{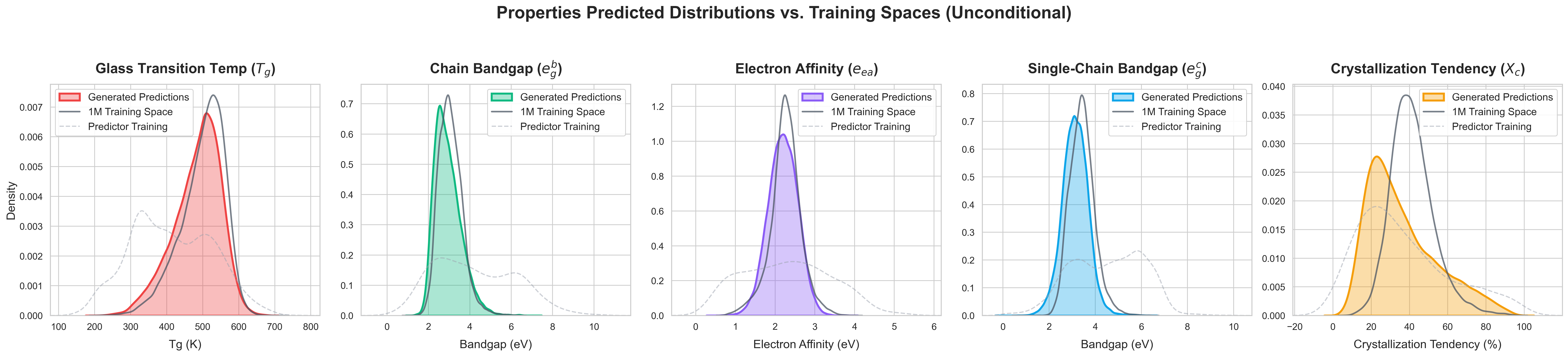}
    \caption{Predicted property distributions of the {\em unconditional} PolymerGPT model.}
    \label{fig:uncond_fig}
    \vspace{-0.1in}
\end{figure*}

We first evaluate PolymerGPT in the unconditional setting, where no property prefix is provided, and the model is trained to generate polymers solely from the pSMILES context. Remarkably, 99.26\% of the generated strings are valid pSMILES, 99.6\% of the valid polymers are unique after canonicalization, and 99.5\% of unique polymers are novel with respect to the training corpus, yielding 98.37\% of samples that simultaneously pass all three filters. Using the pSMILES canonicalization provided by PolyBERT as an additional validity check \cite{polyBERT}, the unconditionally generated polymers also achieved a validity of 98.89\%. {\em The high validity significantly improves existing unconditional generation methods}: polyG2G of 93\% \cite{PolyG2G}, SD-VAE of 13-27\% \cite{batra2020polymers}, IGGM of 44\% \cite{liu2023high}, Mole. Chef of 16.1-89.4\% \cite{kim2023open}, as reported in Table 1 of \cite{POLYTAO}. In particular, PolyG2G, SD-VAE, IGGM, and Mole. Chef have training data size (13K, 250K, 250K, 120K) and generation size (58K, 1K, 10K, 30K, respectively). Due to different training datasets and the number of sampled structures, these results should be treated as literature-based comparisons rather than controlled head-to-head benchmarks. Our high validity matches with the progress in small molecule design, despite the higher complexity in pSMILES strings.

To analyze the property distributions of unconditional samples, we pass generated polymers through TransPolymer and compare the resulting distributions to those of the training data in {\bf Figure~\ref{fig:uncond_fig}}. For most properties, the unconditional model reproduces the broad shape and range of the training distribution, while intrinsically harder properties such as crystallinity $X_c$ show larger discrepancies. The generated distribution for $X_c$ leans toward the predictor’s training distribution, likely because the predictor regresses generated structures toward its sparse, low‑$X_c$‑heavy training support. {\bf Overall, these results indicate that the unconditional PolymerGPT model samples chemically valid, diverse, and novel polymers whose predicted properties span approximately the same regimes as the training data.}

\subsection{Conditional Generation on a Single Property}
\begin{figure}[!htbp]
\centering
\begin{minipage}{0.5\linewidth}
    \centering
    {\fontsize{9}{12}\selectfont
    \setlength{\tabcolsep}{3pt}
    \renewcommand{\arraystretch}{1.05}
    \begin{tabular}{l r r r c r}
    \hline
    Target & VAL & UNQ & NOV & $\hat{T}_g$ & MAE \\
    \hline
    250  & 0.9619 & 0.9754 & 0.9981 & 277.5 $\pm$ 33.6 & 34.8 \\
    350  & 0.9721 & 0.9894 & 0.9983 & 355.9 $\pm$ 40.3 & 31.6 \\
    400  & 0.9739 & 0.9915 & 0.9982 & 401.9 $\pm$ 46.0 & 36.8 \\
    450  & 0.9771 & 0.9939 & 0.9980 & 445.1 $\pm$ 46.2 & 37.4 \\
    500  & 0.9810 & 0.9950 & 0.9979 & 486.6 $\pm$ 43.3 & 35.6 \\
    600  & 0.9774 & 0.9922 & 0.9942 & 553.0 $\pm$ 36.3 & 50.7 \\
    \hline
    \end{tabular}
    } % end 9pt group
    \captionof{table}{$T_g$ (K) sweep generation quality for different target values (Medium model, conditioned on $T_g$ only).}
    \label{tab:tg_sweep_medium}
\end{minipage}\hfill
\begin{minipage}{0.5\linewidth}
    \centering
    \includegraphics[width=\linewidth]{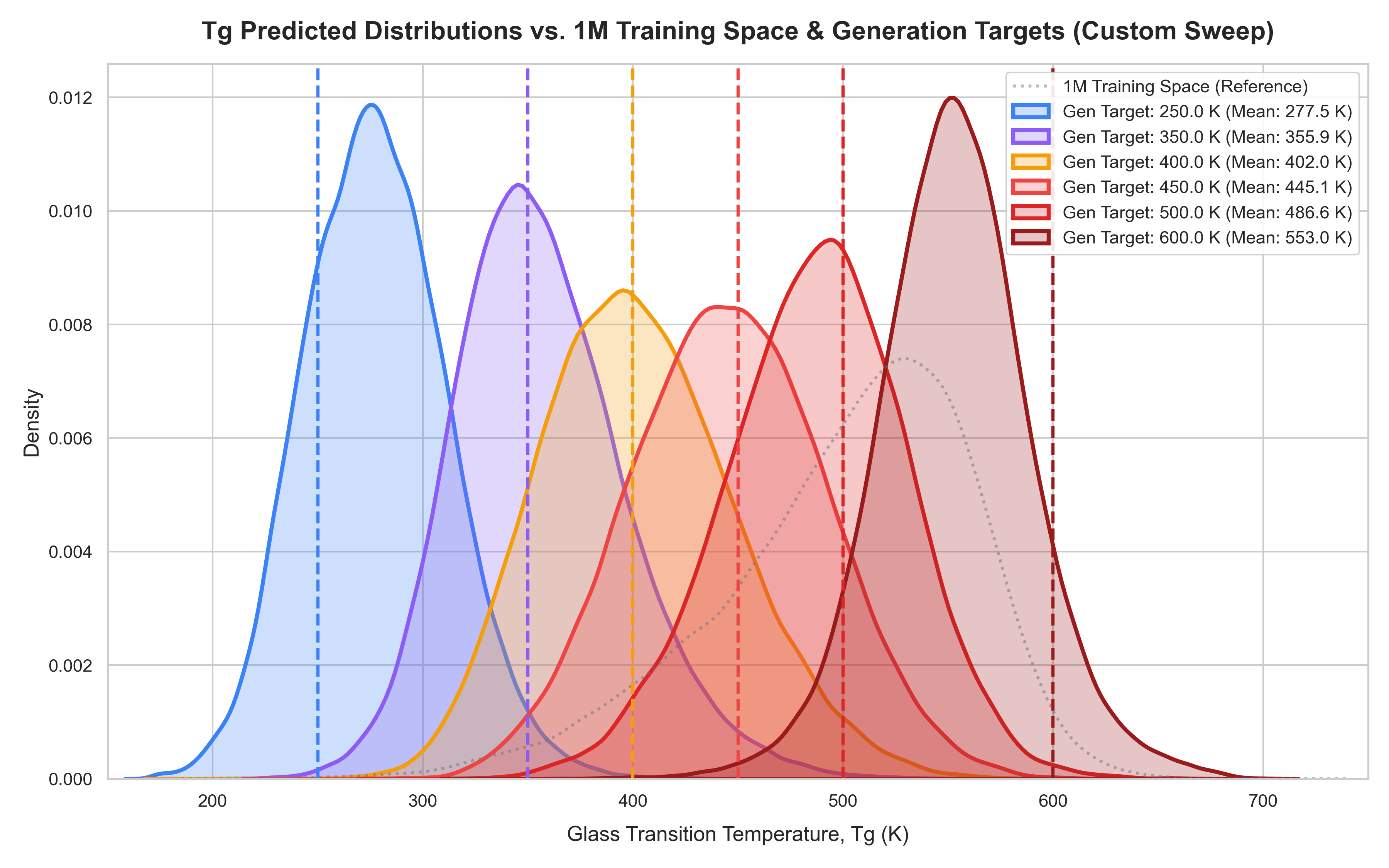}
    \caption{$T_g$ sweep property predictions}
    \label{fig:tgonly}
\end{minipage}
\end{figure}

In this subsection, we evaluate PolymerGPT in a single-property conditional setting, where the model is trained conditioned solely on the glass-transition temperature $T_g$. {\bf Figure~\ref{fig:tgonly}} visualizes the predicted $T_g$ distributions across 6 target values (n = $\sim$50,000 each) spanning the training range and beyond, and {\bf Table~\ref{tab:tg_sweep_medium}} summarizes generation quality and prediction accuracy each target.

Across all tested targets, validity, uniqueness, and novelty remain high, decreasing slightly near the tails of the training distribution. Prediction accuracy degrades
modestly for the highest target, but is otherwise consistent. Despite this, the model consistently generates chemically valid and novel polymers at all target values, demonstrating robust out-of-distribution generalization. For the targeted 500\,K value, 97.4\% of molecules passed all 3 filters. This surpasses PolyT5 and PolyBART, which saw values of 80.6\% and 86.7\% respectively using similar filters. 

To provide an initial assessment of synthetic accessibility, we calculate the Ertl–Schuffenhauer
Synthetic Accessibility (SA) scores \cite{ertl2009estimation} for the polymer-valid structures generated at a high $T_g$ target of 600 K. After replacing the repeat-unit attachment points with hydrogen, 48,603 structures could be scored, with a mean SA score of 3.24 ± 0.62 and a median of 3.15. A mean score of 3.24 is relatively low on the SA scale, which runs from 1, indicating easier synthesis, to 10, indicating greater difficulty.

\smallskip
{\noindent \bf Closure Test}
We use real polymers from the training data for the $T_g$ property predictor, excluding those present in the training data for the generator. Out of $\sim$50,000 predicted structures for a target $T_g$ of 250 K, we found two real polymers \texttt{\seqsplit{"*c1ccc(-c2ccc([Si](CCCC)(CCCC)[Si](*)(CCCC)CCCC)s2)s1"}} and \texttt{\seqsplit{"*CC(*)(C)C(=O)OCCOCC"}} that have experimental $T_g$ of 253\,K and 257\,K respectively, very close to the target. This highlights the reliability of PolymerGPT in generating real polymers with an experimentally validated property.

\subsection{Five-Property Design with Property-Conditioned Generation}
\label{sec:allprop}
\begin{figure}[!htbp]
    \centering
    \includegraphics[width=1\linewidth]{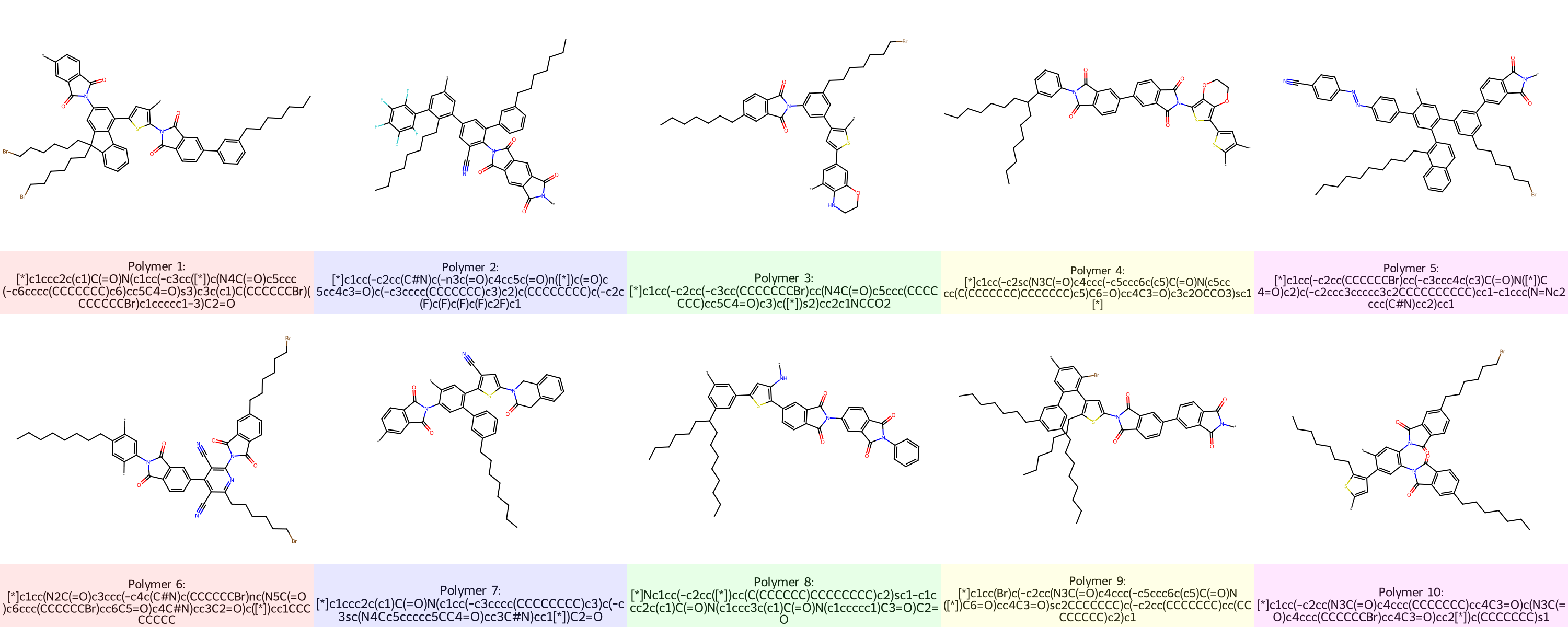}
    \caption{Top-10 generated polymers visualized}
    \label{fig:top10fig}
\end{figure}

We now evaluate PolymerGPT on a five-property design task using the medium model trained on the 1M dataset. We specify targets of $T_g$ = 450\,K for a glass-transition temperature, $e_g^b$ = 2.75\,eV for the bulk electronic band gap, $e_g^c$ = 2.50\,eV for the chain electronic band gap, $e_{\mathrm{ea}}$ = 2.10\,eV for electron affinity, and $X_c$ = 20\% for crystallinity. The remaining 32 property inputs are set to their training-set mean values.
{\bf Figure~\ref{fig:cond_37props} shows the predicted property distributions of generated structures, demonstrating that all 5 properties are closely aligned with the respective target values at the same time.} As we will discuss shortly, this optimization is highly challenging given the limited training samples for the 5 target values. Furthermore, we output the top 10 candidate polymers using a sum of squared errors metric for ranking in {\bf Figure~\ref{fig:top10fig}}. 

{\noindent \bf Selected Polymers} Out of the 32,406 valid polymers, we chose to print out the top 10. Scoring is done by calculating the sum of squared relative errors between the generated properties and their respective target values. This penalty ensures that the selected polymers are well-rounded and do not deviate excessively in any single property. The score is calculated as:
\begin{equation}
    \text{Score} = \sum_{p \in \mathcal{P}} \left( \frac{\hat{y}_p - y_p}{y_p} \right)^2
\end{equation}
where $\mathcal{P} = \{T_g, E_g^b, E_{ea}, E_g^c, X_c\}$ represents the set of desired properties, $\hat{y}_p$ is the predicted property for the generated polymer, and $y_p$ is the corresponding target value. The polymers are then ranked in ascending order by their score, with the lowest score representing the closest match. Table~\ref{tab:top10_molecules} and Figure~\ref{fig:top10fig} display the top ten polymers. As shown in {\bf Table \ref{tab:top10_molecules}}, the top several polymers predicted properties closely align to all target values simultaneously. 

\begin{table}[!htbp]
\centering
{\fontsize{6}{8}\selectfont
\setlength{\tabcolsep}{2pt}
\renewcommand{\arraystretch}{1.05}
\begin{tabular}{p{7cm} r r r r r r}
\hline
pSMILES & $T_g$ (K) & $e_g^b$ (eV) & $e_{\mathrm{ea}}$ (eV) & $e_g^c$ (eV) & $X_c$ & Score \\
\hline
\textit{Target} & 450.0 & 2.750 & 2.100 & 2.500 & 20.000 & --- \\
\hline
{\ttfamily\seqsplit{[*]c1ccc2c(c1)C(=O)N(c1cc(-c3cc([*])c(N4C(=O)c5ccc(-c6cccc(CCCCCCC)c6)cc5C4=O)s3)c3c(c1)C(CCCCCCBr)(CCCCCCBr)c1ccccc1-3)C2=O}}
  & 455.4 & 2.738 & 2.129 & 2.490 & 20.363 & 0.00070 \\
  \hline
{\ttfamily\seqsplit{[*]c1cc(-c2cc(C\#N)c(-n3c(=O)c4cc5c(=O)n([*])c(=O)c5cc4c3=O)c(-c3cccc(CCCCCCC)c3)c2)c(CCCCCCCC)c(-c2c(F)c(F)c(F)c(F)c2F)c1}}
  & 448.1 & 2.780 & 2.060 & 2.492 & 19.684 & 0.00075 \\
  \hline
{\ttfamily\seqsplit{[*]c1cc(-c2cc(-c3cc(CCCCCCCBr)cc(N4C(=O)c5ccc(CCCCCCC)cc5C4=O)c3)c([*])s2)cc2c1NCCO2}}
  & 446.8 & 2.727 & 2.105 & 2.571 & 19.905 & 0.00096 \\
  \hline
{\ttfamily\seqsplit{[*]c1cc(-c2sc(N3C(=O)c4ccc(-c5ccc6c(c5)C(=O)N(c5cccc(C(CCCCCCC)CCCCCCC)c5)C6=O)cc4C3=O)c3c2OCCO3)sc1[*]}}
  & 438.5 & 2.763 & 2.114 & 2.471 & 20.335 & 0.00113 \\
  \hline
{\ttfamily\seqsplit{[*]c1cc(-c2cc(CCCCCCBr)cc(-c3ccc4c(c3)C(=O)N([*])C4=O)c2)c(-c2ccc3ccccc3c2CCCCCCCCCC)cc1-c1ccc(N=Nc2ccc(C\#N)cc2)cc1}}
  & 450.6 & 2.679 & 2.098 & 2.566 & 20.138 & 0.00141 \\
  \hline
{\ttfamily\seqsplit{[*]c1cc(N2C(=O)c3ccc(-c4c(C\#N)c(CCCCCCBr)nc(N5C(=O)c6ccc(CCCCCCBr)cc6C5=O)c4C\#N)cc3C2=O)c([*])cc1CCCCCCCC}}
  & 441.2 & 2.710 & 2.110 & 2.503 & 19.400 & 0.00152 \\
  \hline
{\ttfamily\seqsplit{[*]c1ccc2c(c1)C(=O)N(c1cc(-c3cccc(CCCCCCCC)c3)c(-c3sc(N4Cc5ccccc5CC4=O)cc3C\#N)cc1[*])C2=O}}
  & 449.6 & 2.778 & 2.043 & 2.449 & 20.344 & 0.00154 \\
\hline
{\ttfamily\seqsplit{[*]Nc1cc(-c2cc([*])cc(C(CCCCCC)CCCCCCCC)c2)sc1-c1ccc2c(c1)C(=O)N(c1ccc3c(c1)C(=O)N(c1ccccc1)C3=O)C2=O}}
  & 459.9 & 2.735 & 2.058 & 2.503 & 20.551 & 0.00167 \\
\hline
{\ttfamily\seqsplit{[*]c1cc(Br)c(-c2cc(N3C(=O)c4ccc(-c5ccc6c(c5)C(=O)N([*])C6=O)cc4C3=O)sc2CCCCCCC)c(-c2cc(CCCCCCC)cc(CCCCCCCC)c2)c1}}
  & 450.3 & 2.796 & 2.167 & 2.541 & 20.425 & 0.00201 \\
\hline
{\ttfamily\seqsplit{[*]c1cc(-c2cc(N3C(=O)c4ccc(CCCCCCC)cc4C3=O)c(N3C(=O)c4ccc(CCCCCCBr)cc4C3=O)cc2[*])c(CCCCCCC)s1}}
  & 430.9 & 2.749 & 2.144 & 2.539 & 19.902 & 0.00251 \\
\hline
\end{tabular}
}
\caption{Top-10 generated polymers ranked by score}
\label{tab:top10_molecules}
\end{table}

% Additional discussion and filtered polymers are included within Appendix Sect.~\ref{append_37cond}.

% \begin{table}[!htbp]
% \centering
% {\fontsize{9}{12}\selectfont
% \begin{tabular}{lrrrrr}
% \toprule
% Polymer & $T_g$ & $e_g^b$ & $e_{\mathrm{ea}}$ & $e_g^c$ & $X_c$ \\
% \midrule
% \textit{Target} & 450.0 & 2.750 & 2.100 & 2.500 & 20.000 \\
% Polymer 1       & 455.4 & 2.738 & 2.129 & 2.490 & 20.363 \\
% Polymer 2       & 448.1 & 2.780 & 2.060 & 2.492 & 19.684 \\
% Polymer 3       & 446.8 & 2.727 & 2.105 & 2.571 & 19.905 \\
% \bottomrule
% \end{tabular}
% }
% \caption{Top 3 generated polymers from the 37-property model with the sum of squared relative errors less than 0.001}
% \label{tab:top3_molecules}
% \end{table}

% \begin{figure}
%     \centering
%     \includegraphics[width=.9\linewidth]{figures/top3polymers.png}
%     \caption{Top 3 generated molecules visualized}
%     \label{fig:top3_molecules}
%     \vspace{-0.25in}
% \end{figure}

To assess the benefit of property conditioning, we compared the five-target generator with unconditional and Tg-only generators. Each setting produced $\sim$50,000 samples, and all valid structures were evaluated using the same five TransPolymer property predictors. {\bf Table~\ref{tab:joint_satisfaction}} summarizes the joint target-satisfaction rates. Under strict tolerances of ±10 K for Tg, ±0.25 eV for each electronic property, and ±5 percentage points for Xc, 585 five-target samples satisfied all five criteria, compared with 92 Tg-only and 35 unconditional samples. These correspond to 1.17\%, 0.18\%, and 0.07\% of all generated samples, respectively. Under wider tolerances of ±35 K, ±0.50 eV, and ±10 percentage points, the corresponding rates were 21.11\%, 5\%, and 2.42\%. Although the five-target setting had lower validity, it produced a greater fraction of samples meeting the combined predicted-property criteria. 

\begin{table}[htbp]
\centering
\caption{Joint satisfaction of five target properties for unconditional, $T_g$-only, and five-target generation. Each setting contains $\sim$50,000 generated samples. A joint success is a valid sample whose five TransPolymer-predicted properties fall within the specified tolerances.}
\label{tab:joint_satisfaction}
\begin{tabular}{llrrrr}
\hline
Tolerance & Generation & Valid & Joint & Success / & Success / \\
          & setting    & samples & successes & all (\%) & valid (\%) \\
\hline
Strict & Unconditional & 49,805 & 35  & 0.070 & 0.070 \\
       & $T_g$-only    & 49,026 & 92  & 0.183 & 0.188 \\
       & Five-target   & 32,406 & 585 & 1.166 & 1.805 \\
\hline
Wider  & Unconditional & 49,805 & 1,213  & 2.417  & 2.435 \\
       & $T_g$-only    & 49,026 & 2,508  & 4.998  & 5.116 \\
       & Five-target   & 32,406 & 10,594 & 21.114 & 32.691 \\
\hline
\end{tabular}

\smallskip
\begin{minipage}{0.95\linewidth}
\footnotesize
\textit{Note:} The strict tolerances are $\pm 10$~K for $T_g$,
$\pm 0.25$~eV for each electronic property, and $\pm 5$ percentage
points for $X_c$. The wider tolerances are $\pm 35$~K, $\pm 0.50$~eV,
and $\pm 10$ percentage points, respectively. Success rates are based
on predicted rather than experimentally measured properties.
\end{minipage}
\end{table}

\begin{figure}[!htbp]
    \centering
    \includegraphics[width=.6\linewidth]{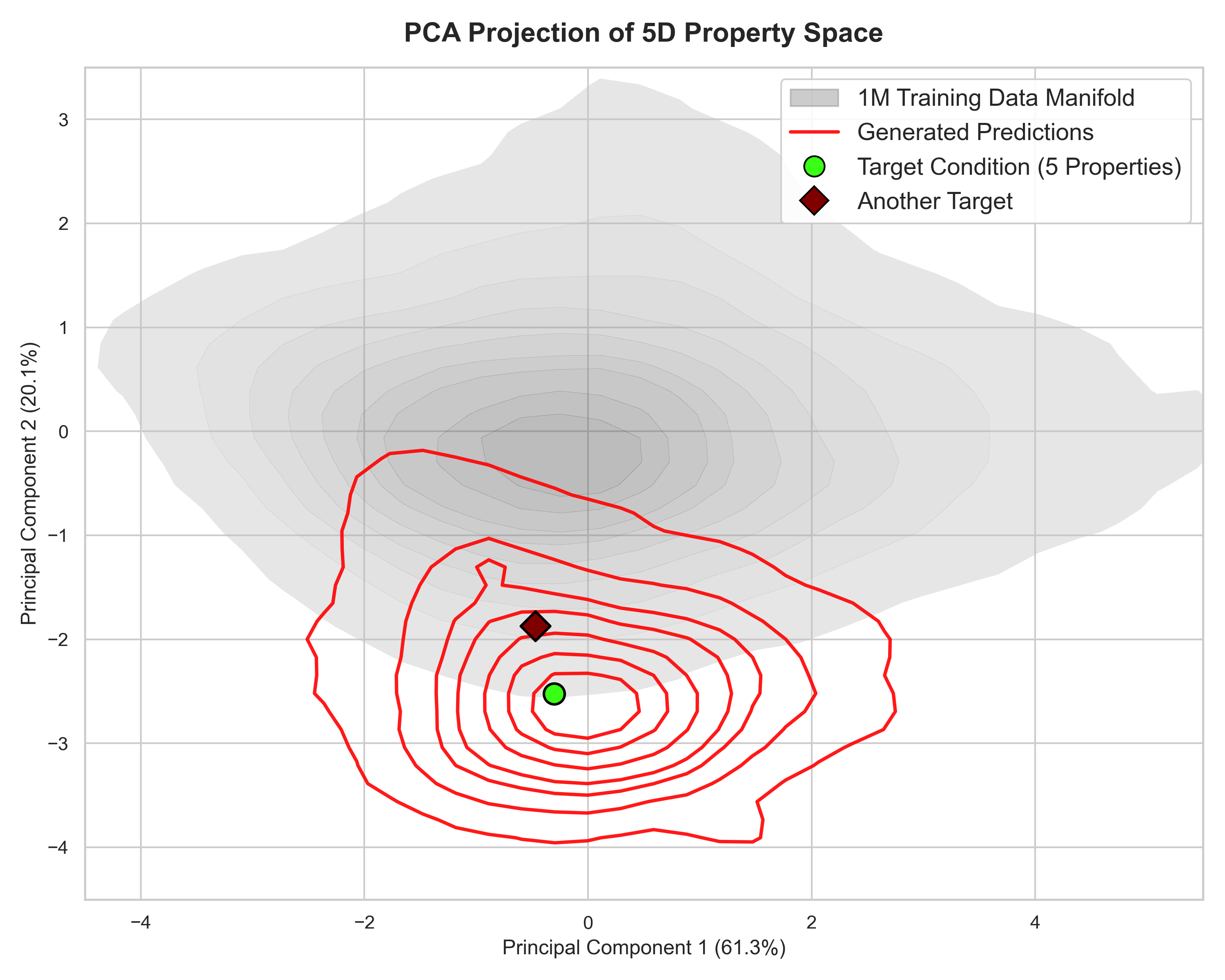}
    \caption{PCA–KDE visualization of the five-property space for generated polymers and the training data. The green circle indicates the specified target.}
    \label{fig:pca_kde}
\end{figure}

\smallskip
We dive deeper into the 5 properties by visualizing a PCA–KDE projection of the 5-property vectors of generated structures vs the training distribution in {\bf Figure~\ref{fig:pca_kde}}. Generated polymers cluster around the specified target region (green circle) and generalize well to the sparsely populated region bordering the training dataset. Generation quality reflects the difficulty of satisfying many coupled constraints at once. The observed validity is 64.58\%, uniqueness 89.68\%, and novelty 99.99\%. The relatively low validity is expected given that the model is being pushed into sparsely explored regions of the joint  property space. Targeting values in more common or covered regions leads to higher validity, as seen in Appendix Sect.~\ref{append_37cond}: targeting the red point closer to the center records a validity of 87\% and targeting the mean of each property achieves a validity of 98.7\%. From an inverse‑design perspective, this tradeoff is acceptable: invalid samples can be discarded cheaply, whereas discovering even a modest number of valid, highly novel polymers that satisfy multiple property constraints is valuable.

\begin{figure}[!htbp]
    \centering
    \includegraphics[width=.5\linewidth]{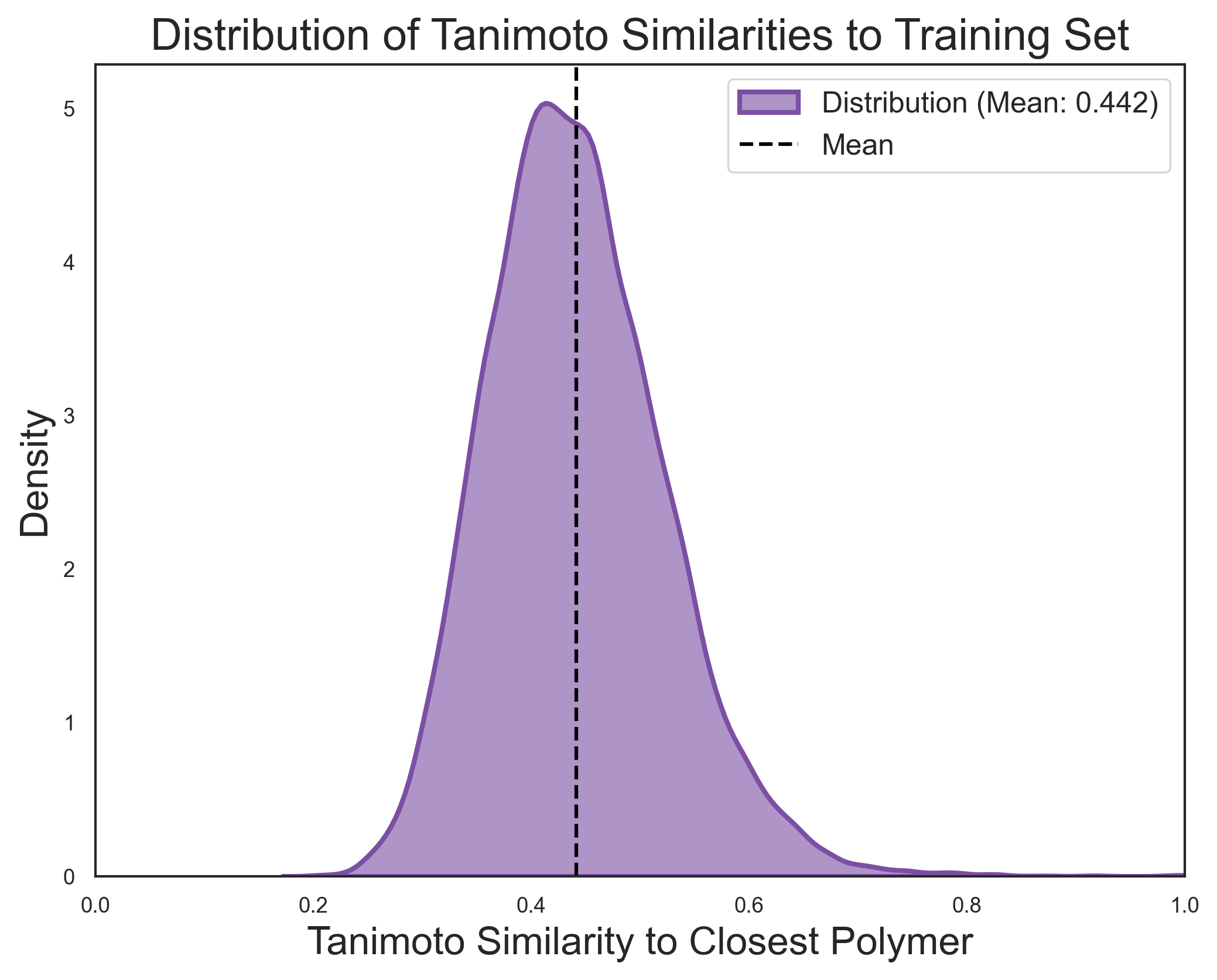}
    \caption{Maximum Tanimoto similarity distribution of generated polymers}
    \label{fig:tanimoto}
\end{figure}

At the same time, uniqueness remains high and novelty is effectively maximal, indicating that the model is exploring genuinely new combinations of properties/structures. We calculated the maximum Tanimoto similarity for all valid generated polymers by comparing  ECFP6 fingerprints against the 1 million training poylmers. {\bf Achieving a mean similarity of 0.44, our results further confirm that PolymerGPT generates structurally distinct polymers rather than memorizing the training data.}  {\bf Figure ~\ref{fig:tanimoto}} displays the full distribution of the maximum tanimoto similarity scores between generated polymers and the training dataset.

\begin{figure}[htbp]
    \centering
    \includegraphics[width=\textwidth]{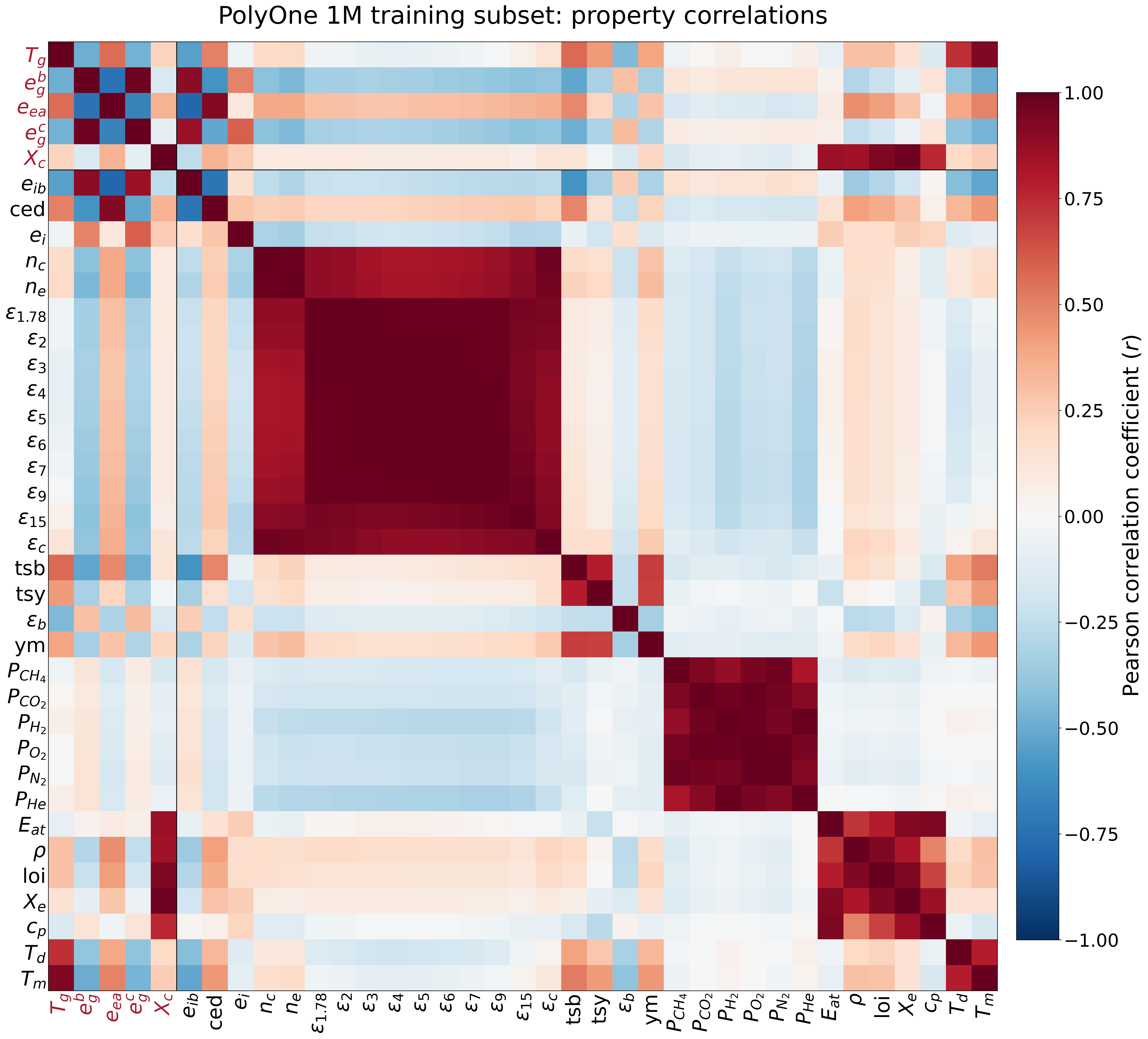}
    \caption{Pearson correlations among the 37 property labels in the PolyOne 1M training subset. The five properties selected for the design demonstration are listed first and
    highlighted in red. Positive and negative correlations are shown in red and blue, respectively. The correlations describe
    model-predicted property labels in the training data.}
    \label{fig:property_correlations}
\end{figure}

To examine relationships among the conditioning properties, we calculated Pearson correlations across the PolyOne 1M training subset  ({\bf Figure~\ref{fig:property_correlations}}). Among the five target properties, the bulk and chain band gaps were strongly correlated ($r$ = 0.976), while electron affinity was negatively correlated with both the bulk band gap ($r$ = -0.737) and the chain band gap ($r$ = -0.667). Several properties held at their training-set means were also strongly correlated with the selected targets. For example, $T_g$ was correlated with melting temperature ($r$ = 0.926), and $X_c$ was correlated with $X_e$ ($r$ = 0.974) and limiting oxygen index ($r$ = 0.933). Thus, assigning mean values to the remaining inputs may influence the generated distribution when those values do not follow the relationships associated with the five selected targets. Investigating how property correlations in the training labels can be exploited to improve conditional generation is an important direction for future work.

\smallskip
{\noindent \bf Impacts of Number of Conditioned Properties}
To assess how the number of conditioned properties affects generation quality and accuracy, we train a series of 1M–scale models conditioned on (i) $T_g$ only, (ii) $T_g$ and $e_g^b$, (iii) $T_g$, $e_g^b$, $T_d$, and (iv) all 37 properties. For each model, we target the mean of the training distribution ($T_g$ = 498.7 K) and sample $\sim$50,000 polymers. {\bf Table~\ref{tab:multi_prop_comparison}} summarizes generation statistics and prediction accuracy.

\begin{table}[!htbp]
\centering
\fontsize{9}{12}\selectfont
\setlength{\tabcolsep}{3pt}
\renewcommand{\arraystretch}{1.05}
\begin{tabular}{l r r r r r}
\hline
N & VAL & UNQ & NOV & $\hat{T}_g$ (K) & MAE (K) \\
\hline
1     & 0.9812 & 0.9943 & 0.9979 & $485.3 \pm 43.0$  & 35.5 \\
2     & 0.9851 & 0.9856 & 0.9967 & $488.4 \pm 42.5$ & 34.2 \\
3     & 0.9835 & 0.9900 & 0.9971 & $490.7 \pm 42.9$ & 34.3 \\
37    & 0.9878 & 0.9652 & 0.9996 & $508.4 \pm 39.3$ & 32.9 \\
\hline
\end{tabular}
\caption{Generation quality and prediction accuracy for models conditioned on 1, 2, 3, and all 37 properties (target: $T_g$ = 498.7 K).}
\label{tab:multi_prop_comparison}
\end{table}

Across the tested configurations, validity and novelty remain consistently high, with the 37-property model achieving the highest validity and novelty. As the number of property inputs increases from one to 37, uniqueness decreases from 0.994 to 0.965, while the predicted $T_g$ MAE decreases from 35.5 K to 32.9 K. The reduction in uniqueness may reflect a more restricted distribution of generated structures under additional conditions. It appears that the model converges to a more restricted chemical space in order to satisfy all the specified property constraints.  We refer to Appendix Sect.~\ref{append_numprops} for plots of the predicted distributions and generation statistics.

\subsection{Scaffold Generation}
\label{sec:scaffold}

\begin{figure}[!htbp]
    \centering
    \includegraphics[width=1\linewidth]{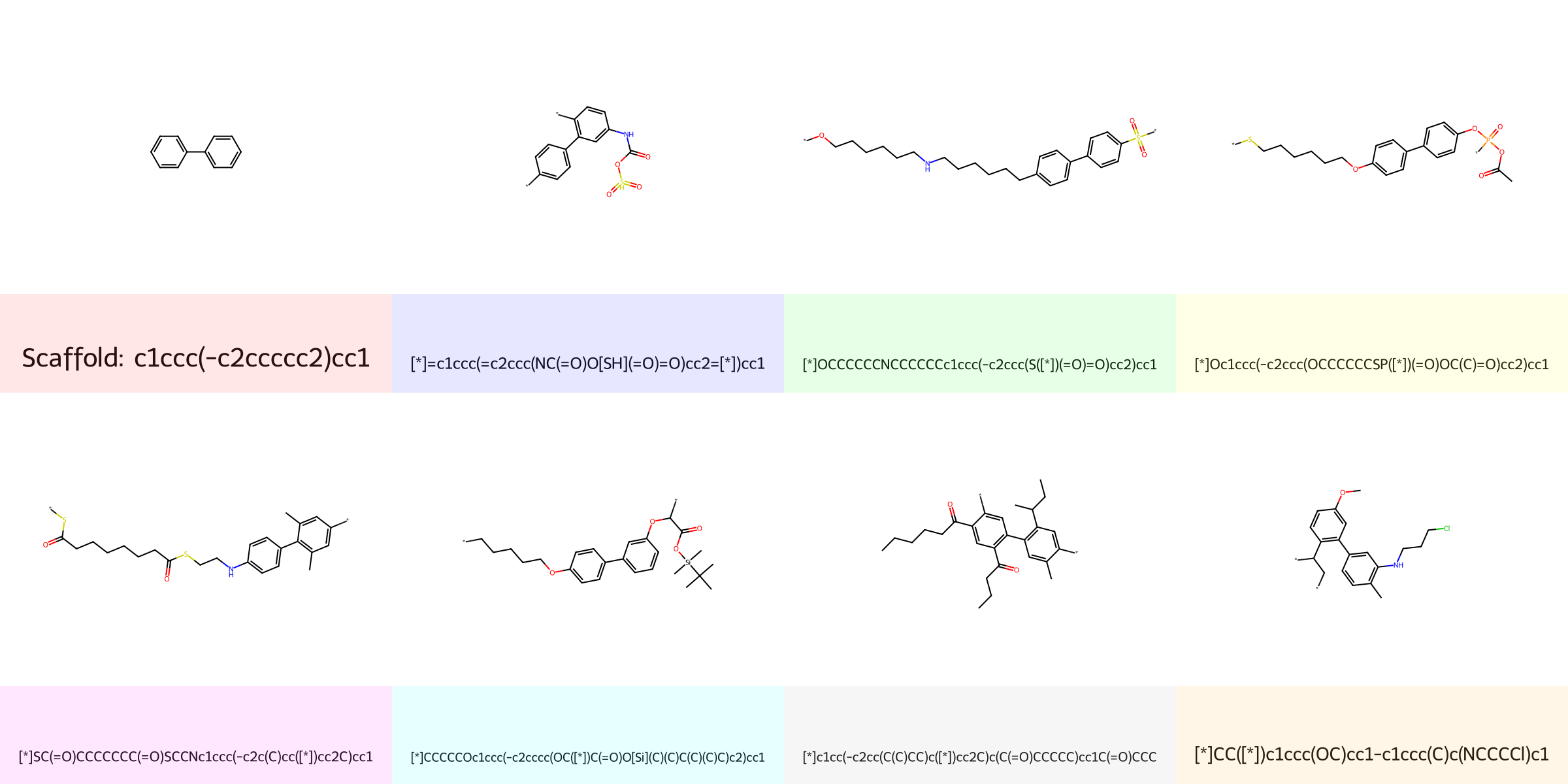}
    \caption{Scaffold-conditioned generation results for the target biphenyl scaffold}
    \label{fig:append_scaff_bip.}
\end{figure}

% {\noindent \bf Unconditional Scaffold Generation} 
We additionally assess scaffold-level control by conditioning PolymerGPT on specific structural motifs. Generating $\sim$50,000 structures using the biphenyl scaffold, \texttt{"c1ccc(-c2ccccc2)cc1"}, the most frequent scaffold in the 1M dataset (6,466 occurrences), we obtain 99.47\% valid pSMILES, 97.18\% uniqueness among valid samples, and 99.73\% novelty relative to the training set, with {\bf 99.81\% of valid molecules containing the desired scaffold}. This illustrates that the model composes diverse, novel structures around the same scaffold while preserving high generation quality. {\bf Figure~\ref{fig:append_scaff_bip.}} displays the scaffold-conditioned generation results. The top-left panel displays the target biphenyl scaffold \\(\texttt{c1ccc(-c2ccccc2)cc1}), which is the most commonly occurring scaffold in the dataset. The remaining panels showcase examples of valid, uniquely generated polymer structures that successfully incorporate the target scaffold along with their pSMILES strings.

\begin{figure}[!htbp]
    \centering
    \includegraphics[width=\linewidth]{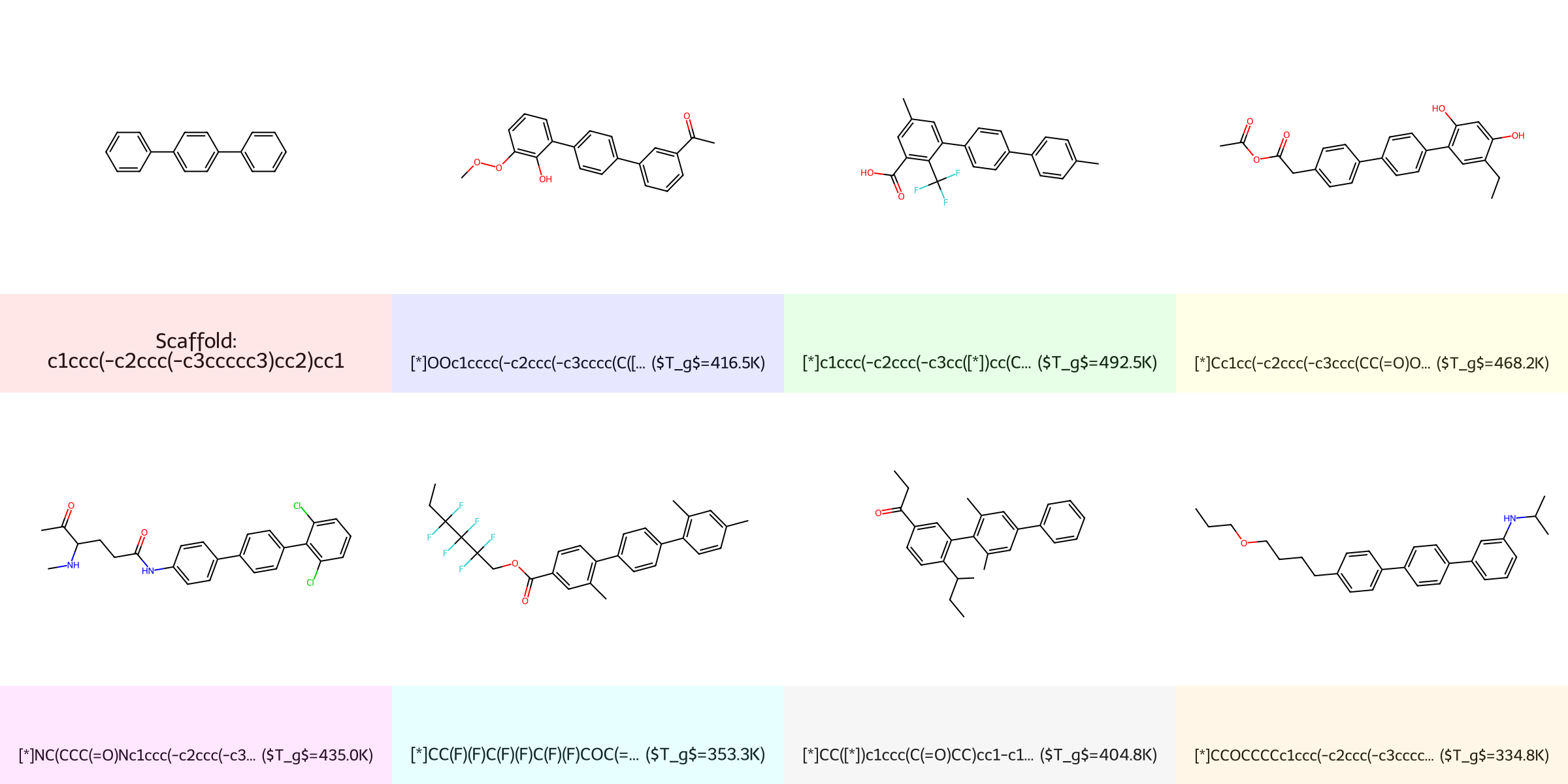}
    \caption{Target $p$-terphenyl scaffold and representative generated polymer structures passing all validation filters.}
    \label{fig:pterphenyl_mols}
\end{figure}

Similarly, we evaluate scaffold-conditioned generation using another common scaffold, the $p$-terphenyl scaffold (\texttt{\seqsplit{c1ccc(-c2ccc(-c3ccccc3)cc2)cc1}}). Out of 50,000 generated structures, the model achieves a validity, uniqueness, and novelty of 99.58\%, 94.51\%, and 100\%, while 99.80\% of valid generations incorporate the target scaffold. {\bf The lower uniqueness relative to unconditional generation reflects the narrower sampling space imposed by a fixed scaffold, not a drop in generation quality.} {\bf Figure~\ref{fig:pterphenyl_mols}} displays the target scaffold alongside representative generated polymer structures passing all filtration steps.

% Figure~\ref{fig:pterphenyl_dist} shows the predicted $T_g$ values of the generated structures (mean 405.53~K, std 55.45~K) compared to the 1M training subset containing $p$-terphenyl scaffolds (mean 443.48~K, std 55.82~K). 

\begin{figure}[!htbp]
    \centering
    \includegraphics[width=0.75\linewidth]{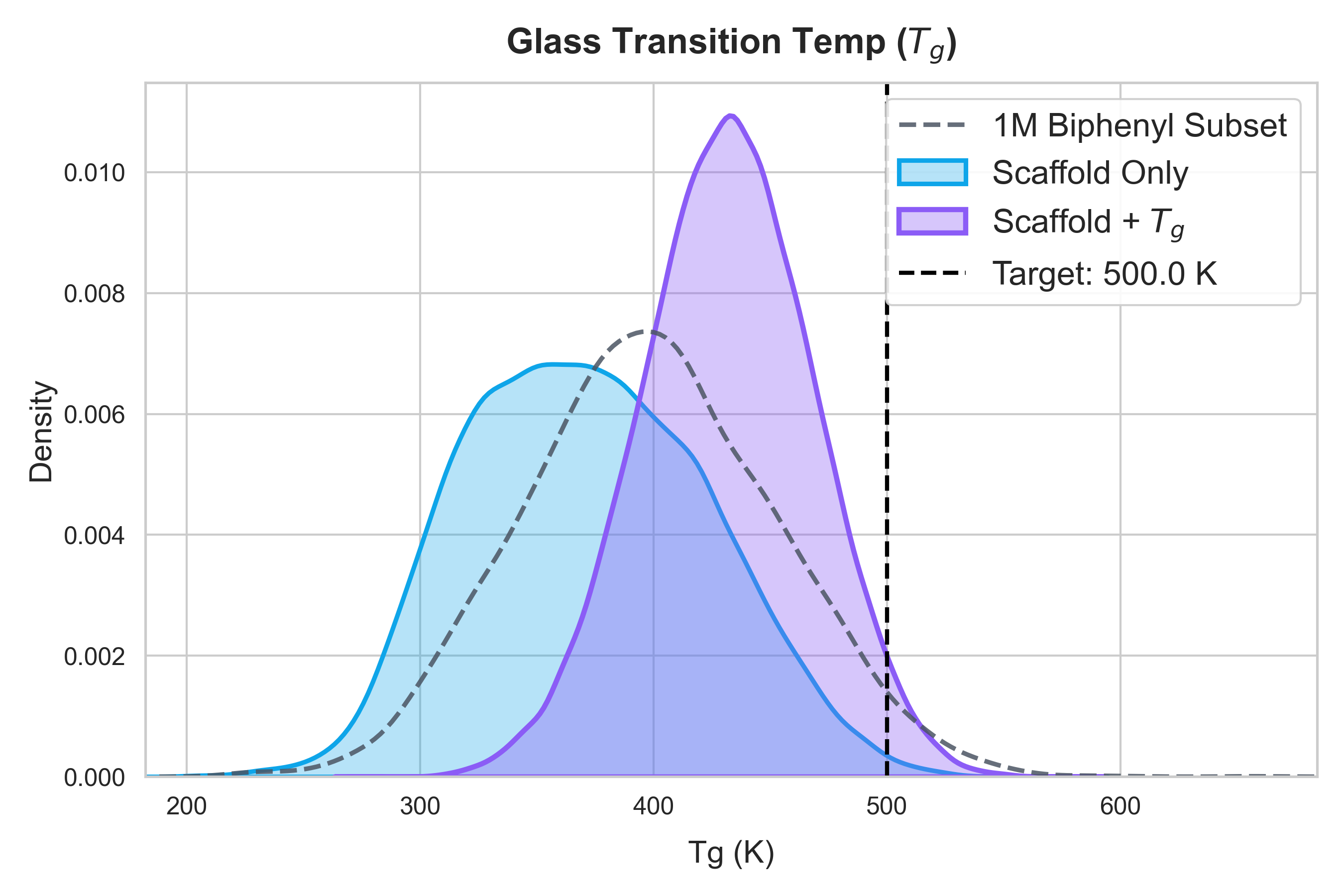}
    \vspace{-0.1in}
    \caption{Predicted properties of structures generated using the biphenyl scaffold}
    \label{fig:scaff_pred}
\end{figure}

\begin{figure}[!htbp]
    \centering
    \includegraphics[width=.75\linewidth]{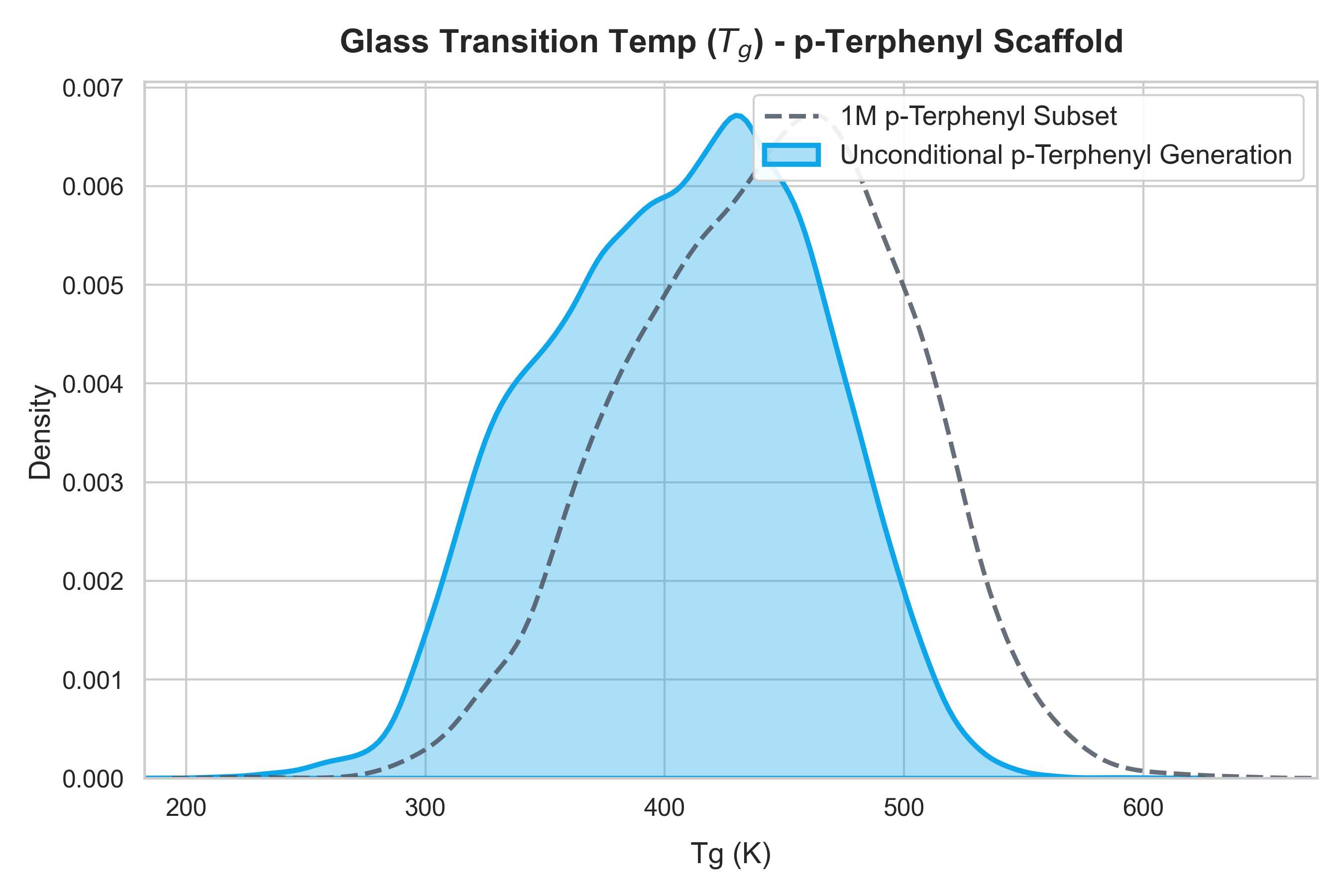}
    \caption{$T_g$ property distribution of $p$-terphenyl scaffold generations compared to the 1M p-Terphenyl subset.}
    \label{fig:pterphenyl_dist}
\end{figure}

% {\noindent \bf Targeted Scaffold Generation}
The unconditional property distribution of scaffold generation for both scaffolds are similar to the distribution of the training data with the scaffolds, as shown in {\bf Figures~\ref{fig:scaff_pred} and \ref{fig:pterphenyl_dist}}. Conditioned on the biphenyl scaffold, we further target a $T_g$ of 500\,K and generate $\sim$50,000 structures. The scaffold condition is still effective, with 99.84\% of valid generations containing the scaffold. The $T_g$ value was purposefully chosen to be an extreme value with respect to biphenyl-containing polymers in the training dataset. As expected and similar to the highest $T_g$ value in Figure~\ref{fig:tgonly}, the model has difficulty in predicting structures of $T_g$ centered at the extreme target, but still successfully shifts the generated distribution upward.

\subsection{Additional Experiments}
\label{sec:extraexp}
\begin{figure}[H]
    \centering
    \includegraphics[width=.75\linewidth]{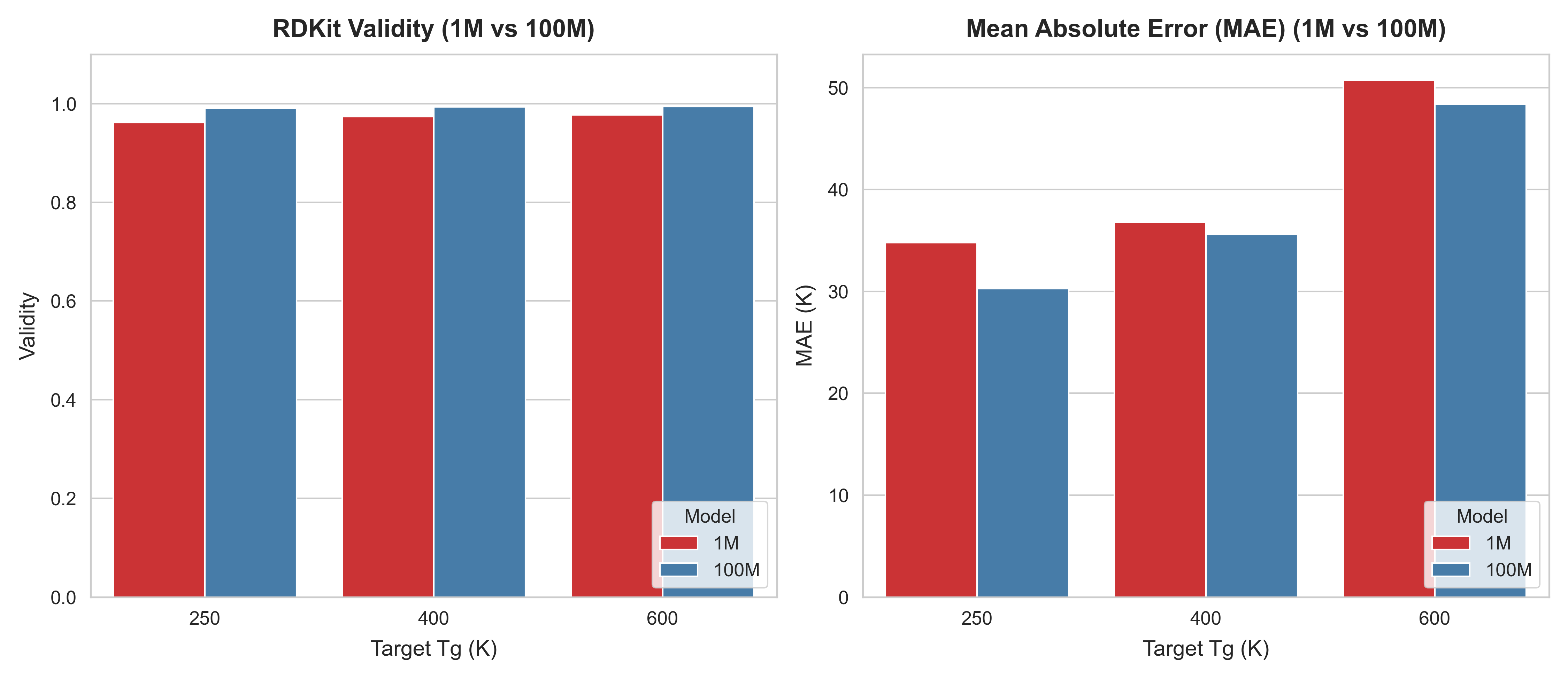}
    \caption{Validity and MAE comparisons across different $T_g$ sweeps for the 1M and 100M models}
    \label{fig:1m100mcompare}
    \vspace{-0.1in}
\end{figure}

{\noindent \bf Impacts of Varied Training size}
We study the effects of training corpus size on conditional generation by fixing the medium model to be conditioned on $T_g$ and sweeping target values across a large range of values. The model is trained on 1M and 100M polymer datasets and selected results are shown in {\bf Figure~\ref{fig:1m100mcompare}}. Across targets, the 100M model achieves higher validity and lower MAE, indicating that larger training data improves the model's generation capabilities. Testing on 6 comprehensive sweeps (in Appendix Sect.~\ref{append_trainsize}) shows consistent results and additional improvement at extreme values. Novelty drops markedly for the 100M model at most targets, which is expected: the 100M training corpus covers a far larger fraction of the pSMILES space, so generated polymers are more likely to already appear in the training set. We choose the 1M dataset as the default training scale for experiments due to training time and compute costs (shown in Appendix Sect.~\ref{append_runtimes}).

\begin{figure}[H]
    \centering
    \includegraphics[width=.75\linewidth]{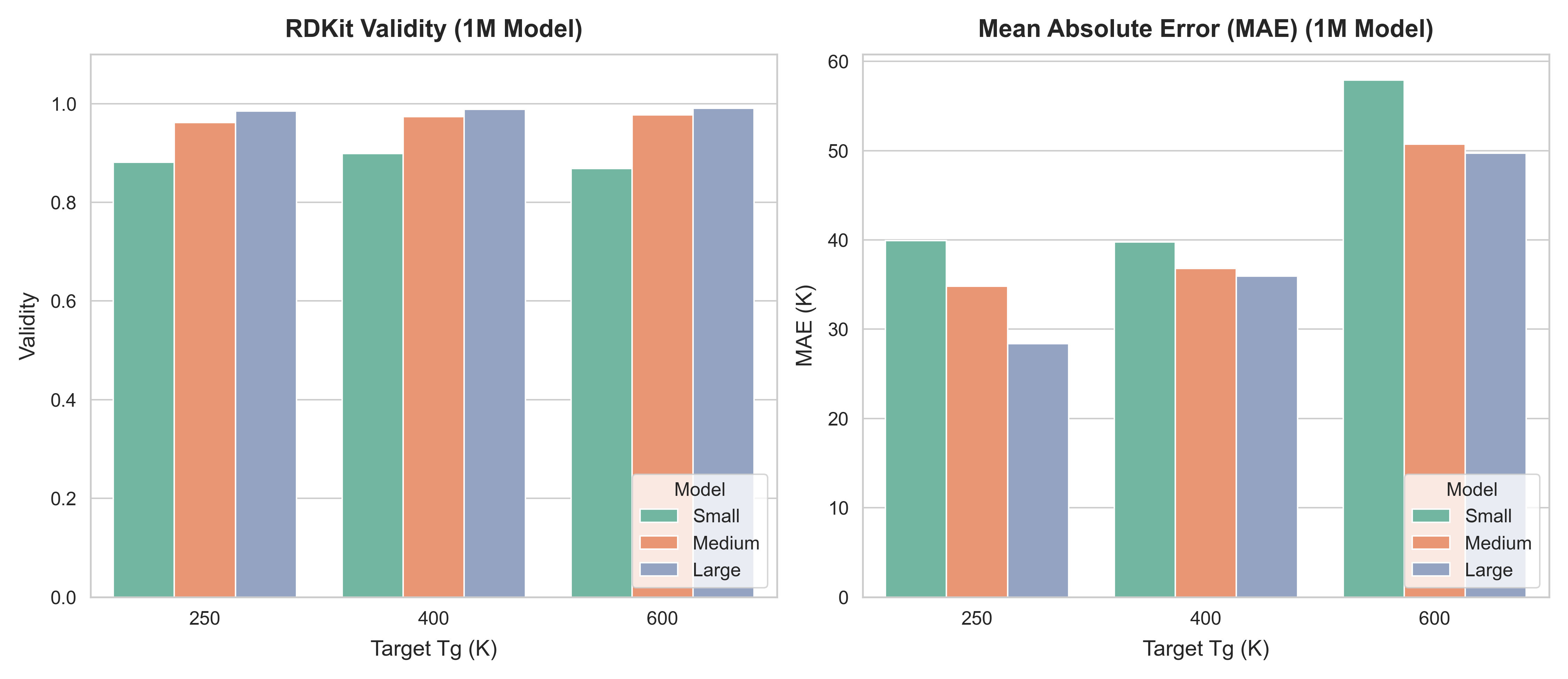}
    \caption{Validity and MAE figures for different model sizes conditioned on $T_g$ and trained on 1M pSMILES}
    \label{fig:modelsize}
\end{figure}

{\noindent \bf Impacts of Varied Model size }
To understand how model capacity affects conditional generation, we compare the small, medium, and large PolymerGPT variants on the 1M training size for a $T_g$ only model in {\bf Figure~\ref{fig:modelsize}}. Across all sweeps, increasing model size consistently improves validity and decreases MAE, most notably in extreme target cases. Further testing (within Appendix Sect.~\ref{append_modelsize}) shows similar results across 6 sweeps, and that uniqueness slightly increases with model size. Testing on the medium and large sizes for the 100M training size (also within Appendix Sect.~\ref{append_modelsize}) is consistent, with the exception of a novelty decrease in the large model.

{\noindent \bf Impacts of Generation size}
We next study how the number of generated samples affects both generation quality and property accuracy. Using the 1M medium model conditioned on $T_g$ only, we sweep four generation sizes (10K, 50K, 100K, and 500K) targeting multiple target sweep values, reporting generation quality metrics in {\bf Figure~\ref{fig:gensize_genstats}} and other outcomes in Appendix Sect.~\ref{append_gensize}. Performance remains highly stable as generation volume increases, and the model reliably centers the distributions around the targets across all scales. This enables large‑scale generation for synthesizability and reactivity screening, without compromising quality or accuracy. 

\begin{figure}[!htbp]
    \centering
    \includegraphics[width=\linewidth]{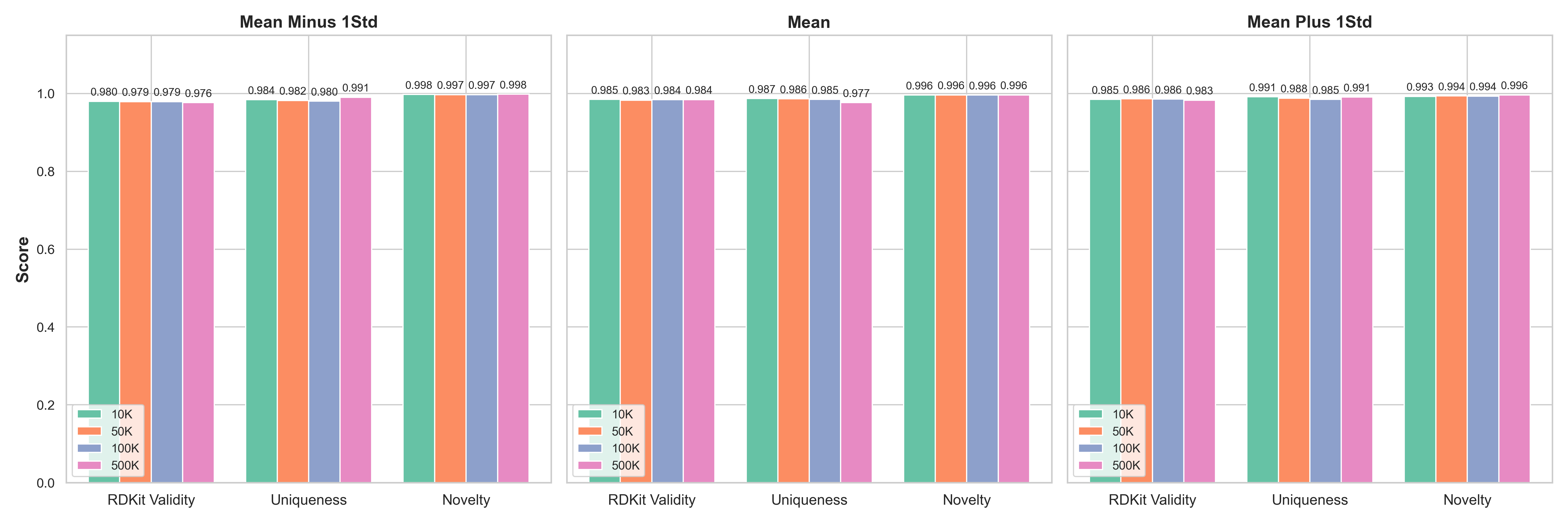}
    \caption{Generation quality metrics (RDKit validity, uniqueness, and novelty) across varying generation sizes (10K to 500K) for three target conditions. Performance remains highly stable as generation volume increases.}
    \label{fig:gensize_genstats}
\end{figure}

\section{Conclusion}
In this work, we presented PolymerGPT, a decoder-only GPT model for generating polymer repeat units under property and scaffold conditions. We evaluated conditional and unconditional generations and examined the effects of model size, training-set size, and generation size. In a five-property design demonstration, selected generated candidates showed close agreement with the specified targets according to TransPolymer predictions. PolymerGPT accepts up to 37 property inputs, providing a basis for future studies of additional target combinations. As the future work, we will study the enhancement of our model with advanced deep generative models. We will also proceed to applications of our model to identify promising candidates for synthesis and experimental validation. Generating candidate polymers together with their promising synthesis routes is also an interesting direction.

\section*{Data and Code Availability}
The training data set PolyOne of 100 million hypothetical polymers is available for academic use at \url{https://doi.org/10.5281/zenodo.7766806}. The glass-transition temperature data set is available at \url{https://www.kaggle.com/code/awguhst/polymer-glass-transition-tg-autogluon/notebook} for fine tuning TransPolymer property prediction. The source code that can be used to reproduce the reported results is temporarily included in the Related Information and will be publicly available at a GitHub repository.

\section*{Supplementary Information}
Comparison of property distributions between the 1M generator training corpus and the TransPolymer
predictor training sets (Figure S1);
Generation quality metrics (validity, uniqueness, novelty) for models conditioned on 1, 2, 3, and all 37 properties (Figure S2);
Property distributions for polymers generated by models conditioned on 1, 2, 3, and all 37 properties (Figure S3);
Accuracy metrics for models conditioned on
1, 2, 3, and all 37 properties (Figure S4);
37 conditional model high validity run on alternate target (Figure S5);
37 conditional model mean target run (Figure S6);
Prediction comparisons across different $T_g$ sweeps for the 1M and 100M models (Figure S7);
Generation statistics comparisons across different $T_g$ sweeps for the 1M and 100M models (Figure S8);
MAE comparisons across different $T_g$ sweeps
for the 1M and 100M models (Figure S9);
Predicted $T_g$ distributions for Small, Medium, and Large models trained on the 1M corpus, across $T_g$
sweep targets (Figure S10);
Generation statistics (validity, uniqueness, novelty) for Small, Medium, and Large models trained on the
1M corpus, across $T_g$ sweep targets (Figure S11);
MAE of predicted $T_g$ for Small, Medium, and
Large models trained on the 1M corpus, across $T_g$ sweep
targets (Figure S12);
Predicted $T_g$ distributions for Medium and Large models trained on the 100M corpus, across $T_g$ sweep
targets (Figure S13);
Generation statistics (validity, uniqueness, and novelty) for Medium and Large models trained on the
100M corpus, across $T_g$ sweep targets (Figure S14);
Mean absolute error (MAE) of predicted $T_g$
for Medium and Large models trained on the 100M corpus, across $T_g$ sweep targets (Figure S15);
Predicted $T_g$ distributions for generation sizes ranging from 10K to 500K (Figure S16);
Mean Absolute Error (MAE) of predicted $T_g$
values across different generation sizes (Figure S17);

Five conditional high validity run generation metrics for alternate target (Table S1);
Property prediction accuracy for the 5 property conditional model on alternate target (Table S2);
Five conditional mean target generation metrics (Table S3);
Property prediction accuracy for the 5 property model’s mean target run (Table S4);
$T_g$ (K) sweep generation quality: 1M vs. 100M
training (Table S5);
$T_g$ (K) sweep generation quality by model size
and target condition (1M training) (Table S6);
$T_g$ (K) sweep generation quality for Medium
and Large models (100M training) (Table S7);
$T_g$ sweep generation quality and prediction ac-
curacy across generation sizes (1M model, conditioned on
$T_g$ only) (Table S8);
Training times for PolymerGPT generator models (Table S9);
Training times for TransPolymer property predictor models fine-tuned on each of the five target properties (Table S10);
Generation times and throughput (structures per second) for PolymerGPT (Table S11);
Prediction times and throughput for TransPolymer (Table S12);
Learning-rate configuration by model variant (Table S13);
Three model variants and their detailed settings (Table S14);
Dataset information for the property prediction model (Table S15);
Hyperparameters and augmentation limit per property symbol (Table S16);
Hyperparameter search space used for TransPolymer fine-tuning (Table S17) (PDF).

% The source code of the developed software (ZIP).

\section*{Author Contributions}
CP performed method development, computational experiments, result interpretation, visualization, and writing.  AG performed conceptualization, method development, computational experiments, result interpretation, and writing. AF contributed through conceptualization, funding acquisition, result interpretation, and writing—review \& editing. ZJ contributed through conceptualization and writing—review \& editing. YL contributed through experimental design and writing—review \& editing.  CZ provided conceptualization, investigation, resources, funding acquisition, project administration, supervision and writing.

\section*{Competing interests}
All authors declare no financial or non-financial competing interests. 

\bibliographystyle{unsrt}
\bibliography{references}

\clearpage
\appendix
\setcounter{table}{0}
\renewcommand{\thetable}{S\arabic{table}}
\setcounter{figure}{0}
\renewcommand{\thefigure}{S\arabic{figure}}

% \section{Properties}
% We consider 37 polymer properties spanning electronic, mechanical, transport,
% and thermal behavior. Table~\ref{tab:property_symbols} lists each property,
% its symbol, and unit, including electronic gaps and ionisation energies,
% dielectric constants, tensile and elastic moduli, gas permeabilities, density
% and crystallinity measures, heat capacity, and key thermal transition
% temperatures ($T_d$, $T_g$, $T_m$).

\newpage
\section{Generation}

\subsection{Generator vs.\ Predictor Distributions}

\begin{figure}[!htbp]
    \centering
    \includegraphics[width=\linewidth]{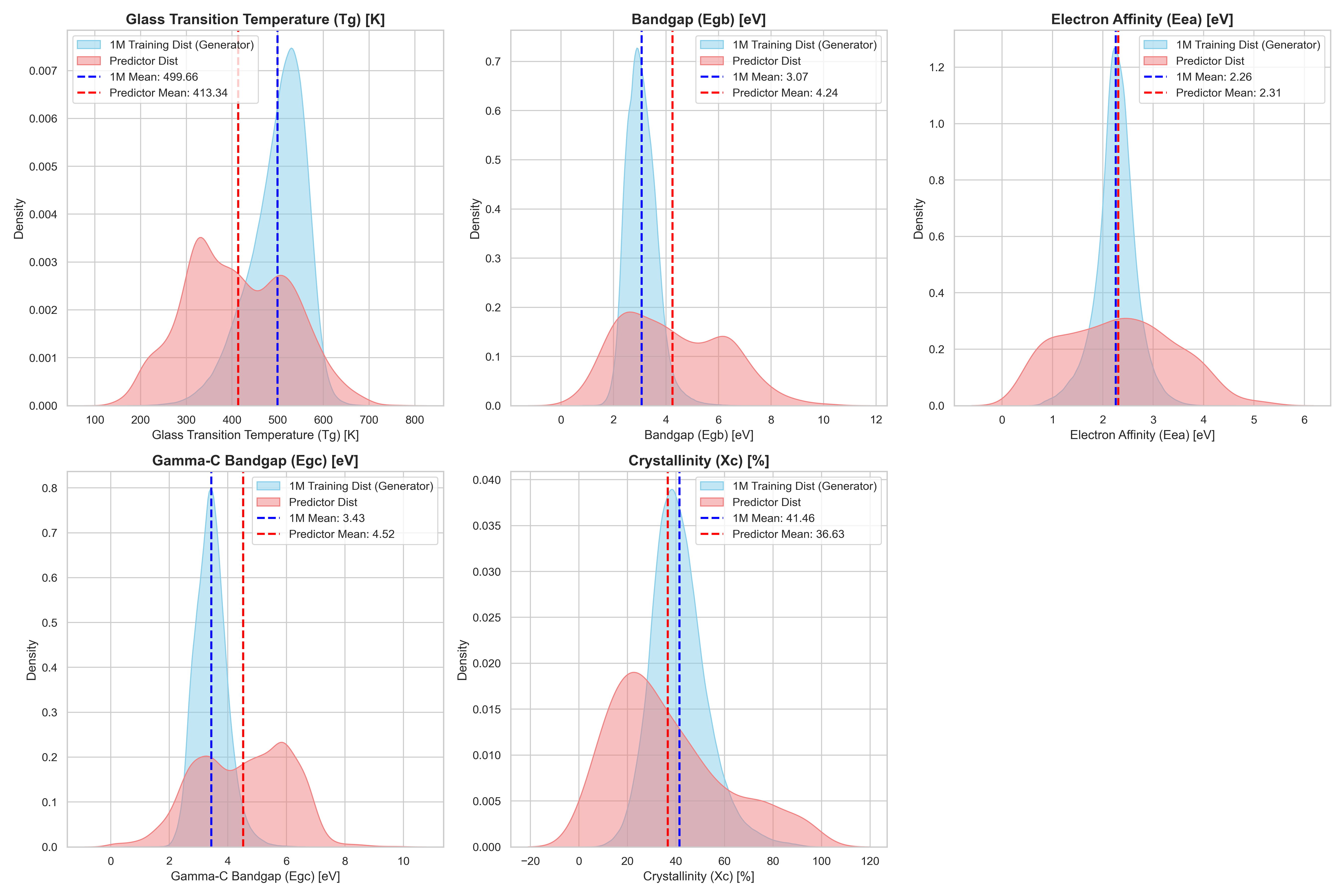}
    \caption{Comparison of property distributions between the 1M generator
    training corpus and the TransPolymer predictor training sets.}
    \label{fig:gen_vs_pred_dist}
\end{figure}

Figure~\ref{fig:gen_vs_pred_dist} compares the property distributions of the 1M generator training set against those of the TransPolymer predictor training sets. The two differ in both origin and coverage: the predictor datasets are derived from experimental measurements and DFT calculations under strict curation criteria, while the generator dataset consists of hypothetically enumerated polymer structures. Their distributional mismatch can introduce a systematic bias when the generator is conditioned on target values that lie outside the predictor's training support, as the predictor may extrapolate unreliably in those regions. To mitigate this, we focus our conditional generation experiments mostly on target values that fall within the overlapping support of both distributions, ensuring generated structures remain within a regime where the predictor provides reliable supervision.

\clearpage

\subsection{Impacts of Number of Properties Conditioned}\label{append_numprops}
\begin{figure}[!htbp]
    \centering

    \includegraphics[width=\linewidth]{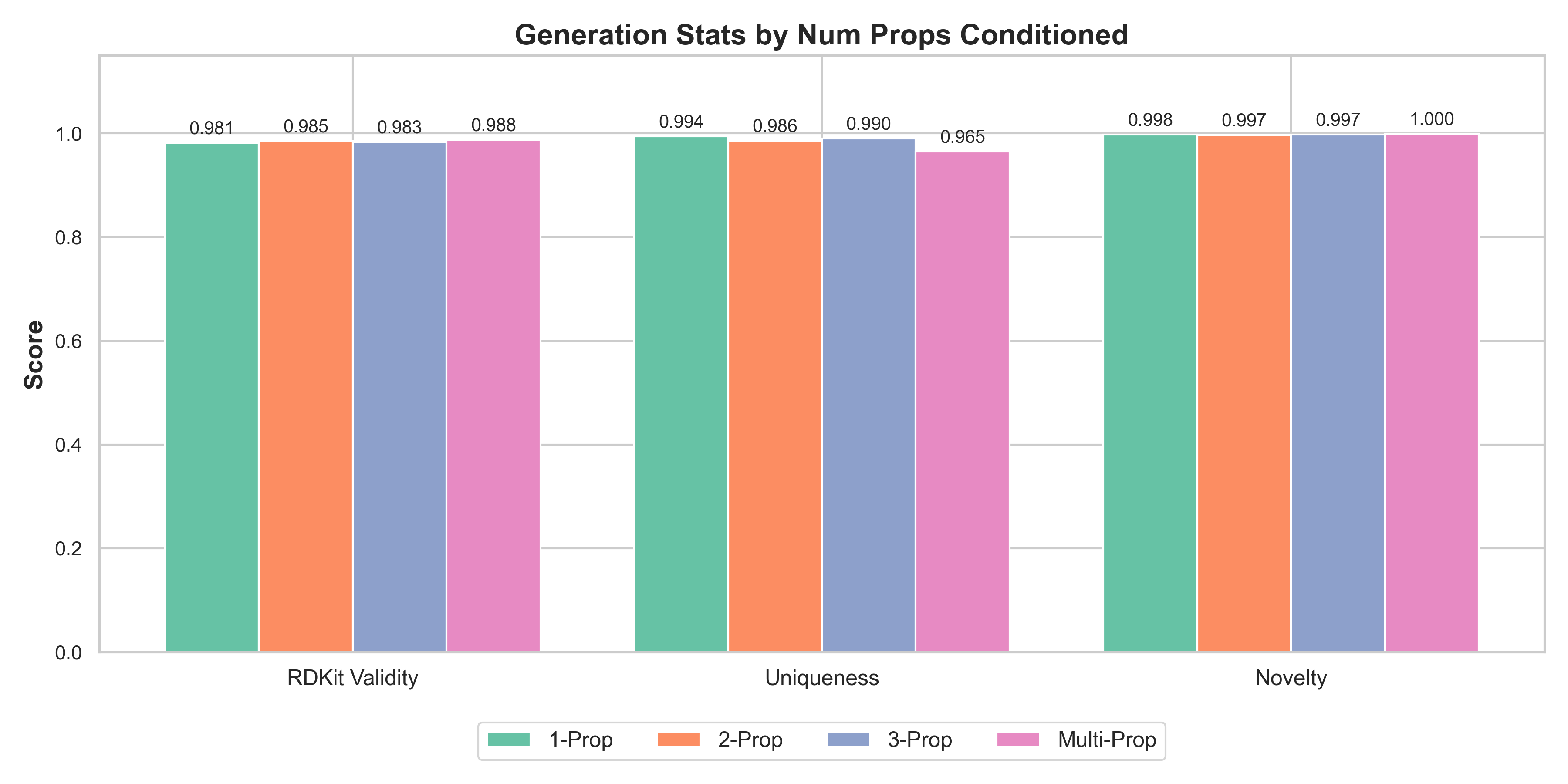}
        \caption{Generation quality metrics (validity, uniqueness, novelty) for models conditioned on 1, 2, 3, and all 37 properties.}
        \label{fig:numprops_genstats}

    \vspace{0.5em}

    \begin{minipage}[t]{0.48\linewidth}
        \centering
        \includegraphics[width=1\linewidth]{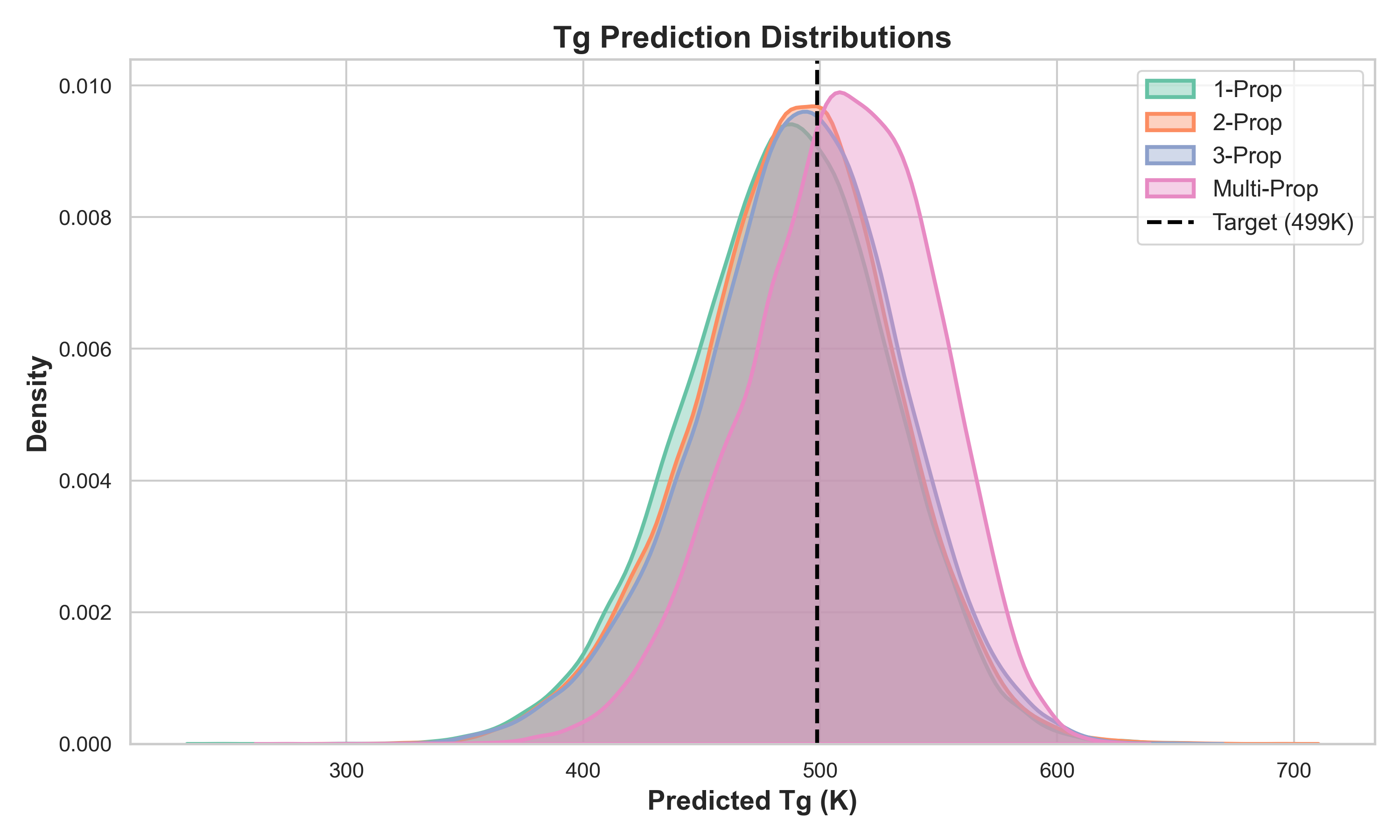}
    \caption{Property distributions for polymers generated by models conditioned on 1, 2, 3, and all 37 properties.}
    \label{fig:numprops_compare}
        
    \end{minipage}\hfill
    \begin{minipage}[t]{0.48\linewidth}
        \centering
        \includegraphics[width=.8\linewidth]{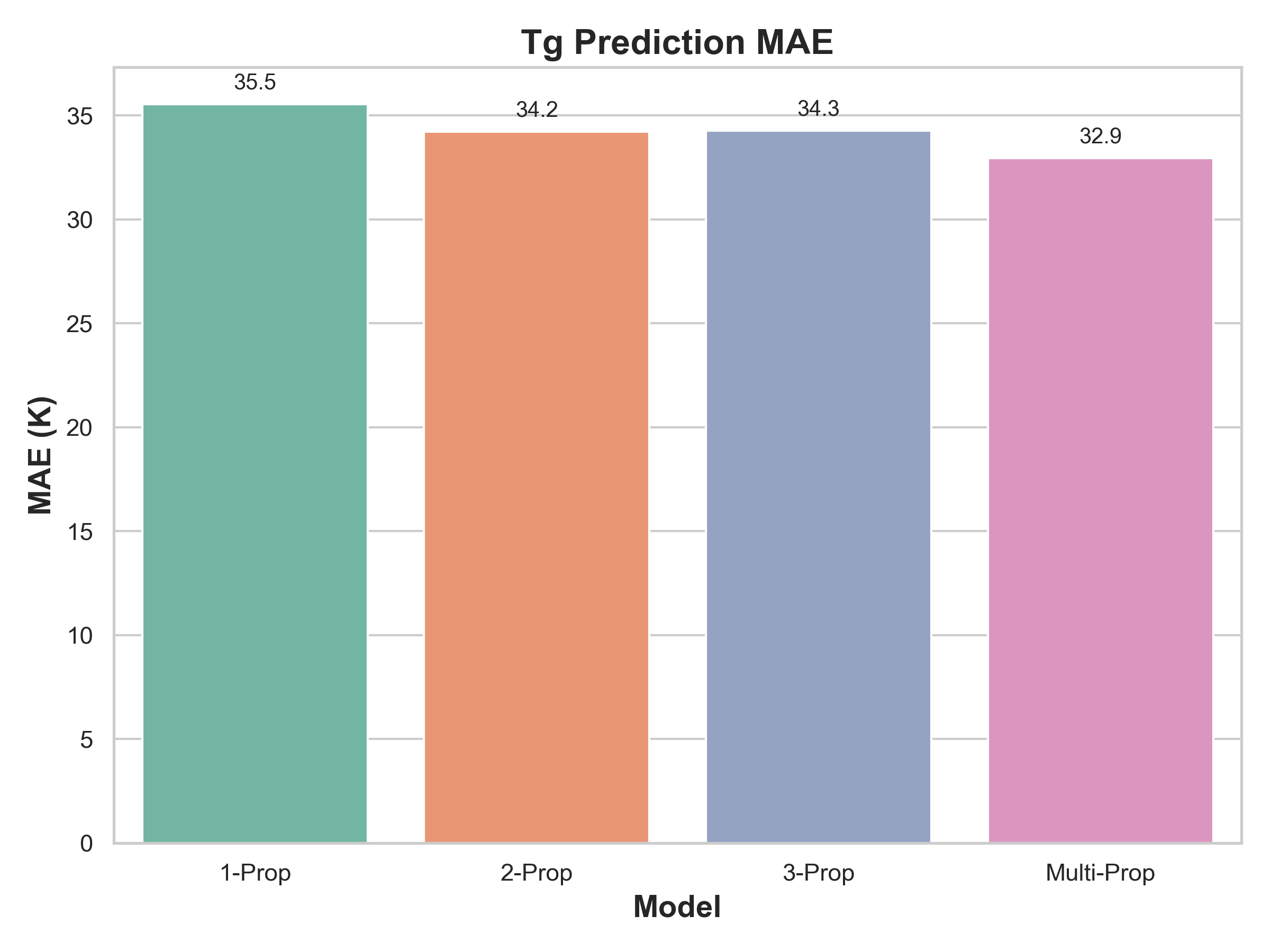}
        \caption{Accuracy metrics for models conditioned on 1, 2, 3, and all 37 properties.}
        \label{fig:numprops_predacc}
    \end{minipage}

\end{figure}

Figure~\ref{fig:numprops_genstats} summarizes generation statistics (validity, uniqueness, and novelty) by models conditioned on different numbers of properties (1, 2, 3, and all 37). Figure~\ref{fig:numprops_compare} and Figure~\ref{fig:numprops_compare} show the distributions of predicted $T_g$ values for generated polymers and MAE to the targeted $T_g$ values, respectively. All statistics remain consistent except for minor decreases in uniqueness and MAE as the number of properties conditioned on increases

\clearpage

\subsection{Five Property Conditional Model}\label{append_37cond}

\begin{figure}[!htbp]
    \centering
    \includegraphics[width=1\linewidth]{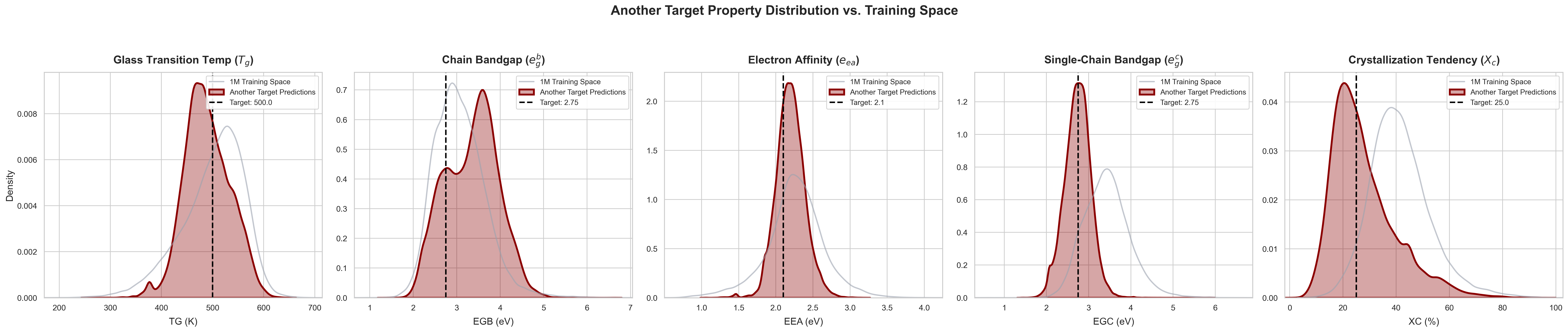}
    \caption{Five conditional model high validity run on alternate target.}
    \label{fig:supp_alt_target}
\end{figure}

\textbf{High Validity Run (another target) } In Figure~\ref{fig:supp_alt_target}, we display a run for the 5 conditional model in which there is high validity, corresponding to the another target label in the main paper's KDE-PCA plot. Most property values are reasonably aligned with their targets, though deviations are observed for $e_g^b$ and $X_c$. Table~\ref{tab:supp_alt_gen_metrics} and Table~\ref{tab:supp_alt_prop_accuracy} show the generation and accuracy prediction.

\begin{table}[!htbp]
\centering
\fontsize{10}{12}\selectfont
\setlength{\tabcolsep}{3pt}
\renewcommand{\arraystretch}{1.05}
\begin{tabular}{r r r r r r r}
\hline
Total & Valid & VAL & Unique & UNQ & Novel & NOV \\
\hline
50{,}176 & 43{,}666 & 0.8703 & 38{,}337 & 0.8780 & 38{,}334 & 0.9999 \\
\hline
\end{tabular}
\caption{Five conditional high validity run generation metrics for alternate target.}
\label{tab:supp_alt_gen_metrics}
\end{table}

\begin{table}[!htbp]
\centering
\fontsize{10}{12}\selectfont
\setlength{\tabcolsep}{3pt}
\renewcommand{\arraystretch}{1.05}
\begin{tabular}{l r r r r}
\hline
Property & Target & Mean & Std & MAE \\
\hline
$T_g$ (K)              & 500.0 & 488.12 & 44.41 & 37.35 \\
$e_g^b$ (eV)           & 2.75  & 3.335  & 0.599 & 0.696 \\
$e_{\mathrm{ea}}$ (eV) & 2.10  & 2.191  & 0.193 & 0.166 \\
$e_g^c$ (eV)           & 2.75  & 2.737  & 0.308 & 0.242 \\
$X_c$                  & 25.0  & 27.89  & 12.10 & 9.08 \\
\hline
\end{tabular}
\caption{Property prediction accuracy for the 5 property conditional model on alternate target.}
\label{tab:supp_alt_prop_accuracy}
\end{table}

\clearpage

\begin{figure}[!htbp]
    \centering
    \includegraphics[width=1\linewidth]{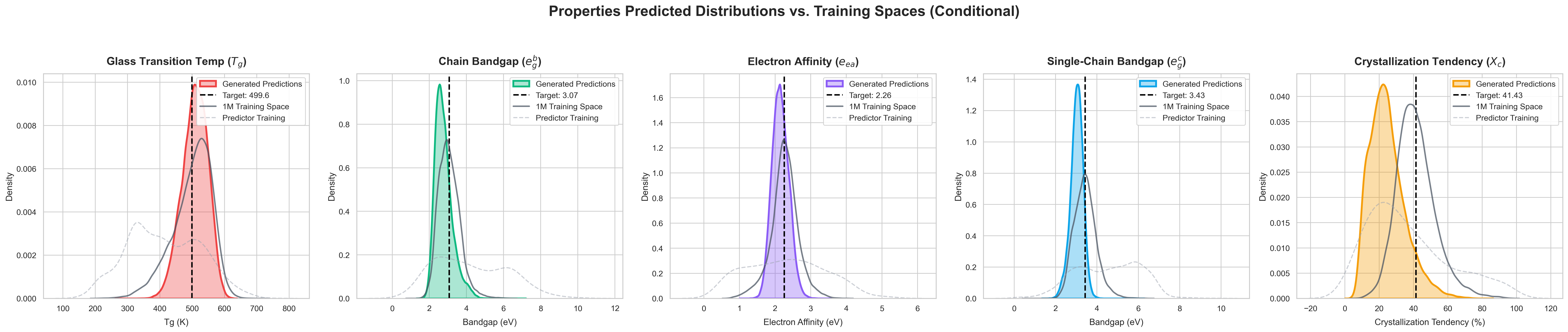}
    \caption{Five conditional model mean target run}
    \label{fig:supp37hgihval}
\end{figure}

\textbf{High Validity Run (mean target) } In Figure~\ref{fig:supp37hgihval}, we display a run for the 5 conditional model targeting the mean values of the training distribution. All property values are aligned well, with the exception of $X_c$. Table~\ref{tab:supp_37_gen_metrics} and Table~\ref{tab:supp_37_prop_accuracy} show the generation and accuracy prediction.

\begin{table}[!htbp]
\centering
\fontsize{10}{12}\selectfont
\setlength{\tabcolsep}{3pt}
\renewcommand{\arraystretch}{1.05}
\begin{tabular}{r r r r r r r}
\hline
Total & Valid & VAL & Unique & UNQ & Novel & NOV \\
\hline
50{,}176 & 49{,}564 & 0.9878 & 47{,}838 & 0.9652 & 47{,}819 & 0.9996 \\
\hline
\end{tabular}
\caption{Five conditional mean target generation metrics.}
\label{tab:supp_37_gen_metrics}
\end{table}

\begin{table}[!htbp]
\centering
\fontsize{10}{12}\selectfont
\setlength{\tabcolsep}{3pt}
\renewcommand{\arraystretch}{1.05}
\begin{tabular}{l r r r r}
\hline
Property & Target & Mean & Std & MAE \\
\hline
$T_g$ (K)              & 499.6 & 508.35 & 39.33 & 32.75 \\
$e_g^b$ (eV)           & 3.07  & 2.775  & 0.477 & 0.475 \\
$e_{\mathrm{ea}}$ (eV) & 2.26  & 2.151  & 0.233 & 0.210 \\
$e_g^c$ (eV)           & 3.43  & 3.026  & 0.308 & 0.426 \\
$X_c$                  & 41.43 & 24.72  & 10.43 & 17.78 \\
\hline
\end{tabular}
\caption{Property prediction accuracy for the 5 property model's mean target run.}
\label{tab:supp_37_prop_accuracy}
\end{table}

\clearpage
\subsection{1M vs 100M Models}\label{append_trainsize}
\begin{figure}[!htbp]
    \centering
    \includegraphics[width=0.95\textwidth]{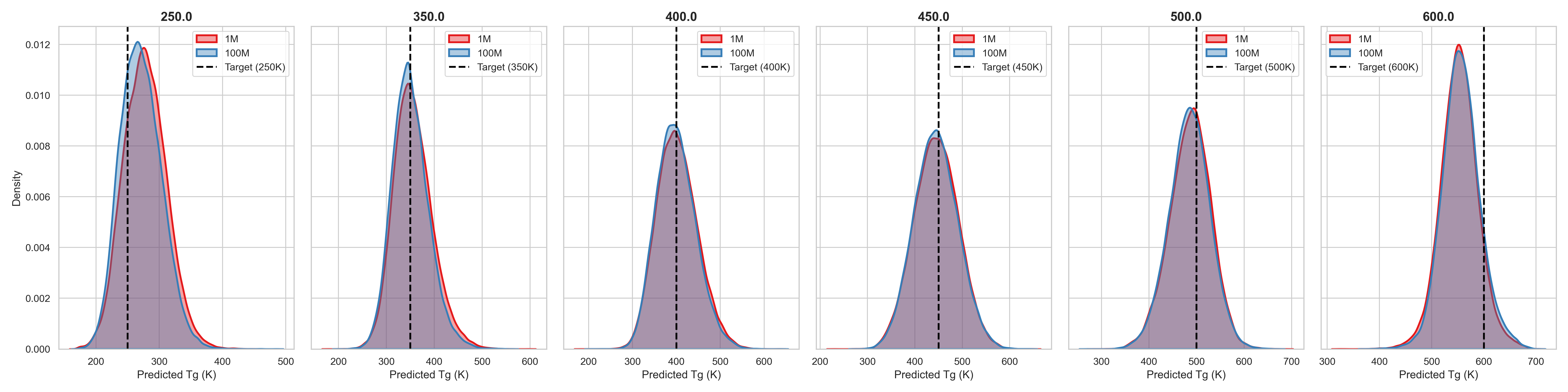}
    \caption{Prediction comparisons across different $T_g$ sweeps for the 1M and 100M models.}
    \label{fig:oneM_vs_hundredM_pred}
\end{figure}

\begin{figure}[!htbp]
    \centering
    \includegraphics[width=0.95\textwidth]{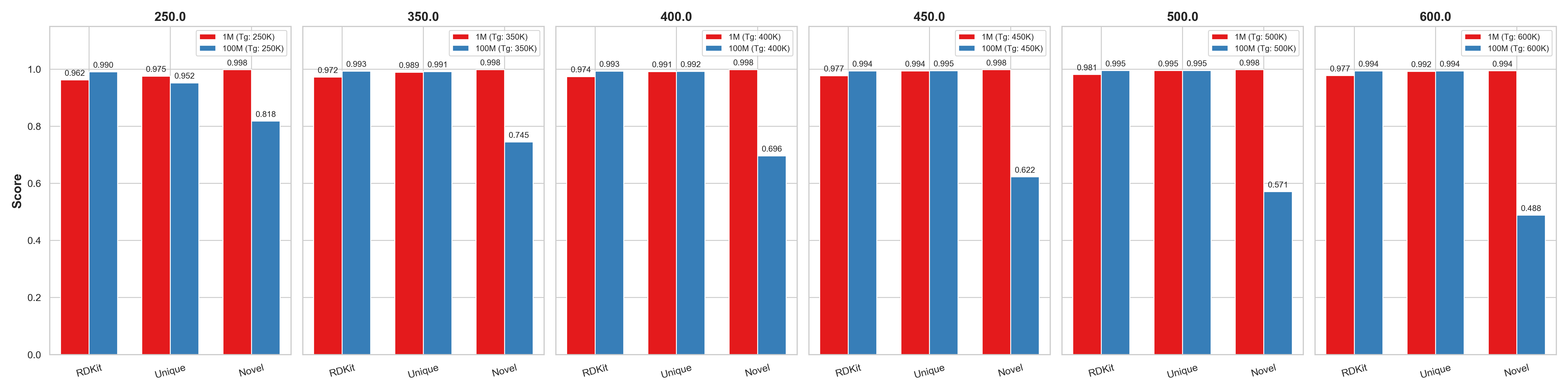}
    \caption{Generation statistics comparisons across different $T_g$ sweeps for the 1M and 100M models.}
    \label{fig:oneM_vs_hundredM_gen}
\end{figure}

\begin{figure}[!htbp]
    \centering
    \begin{minipage}{0.49\textwidth}
        \centering
        \includegraphics[width=1\linewidth]{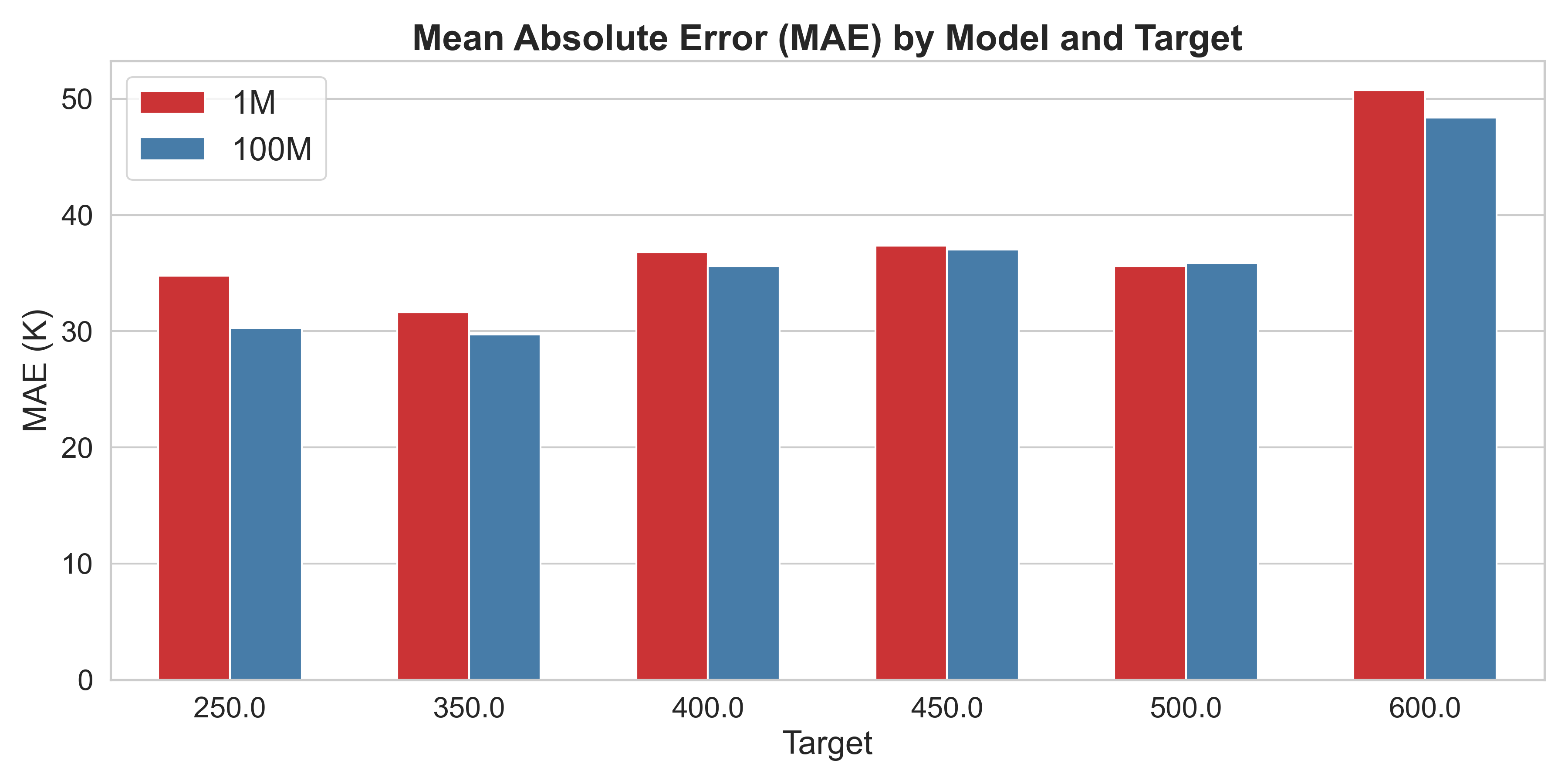}
        \caption{MAE comparisons across different $T_g$ sweeps for the 1M and 100M models.}
        \label{fig:oneM_vs_hundredM_mae}
    \end{minipage}\hfill
    \begin{minipage}{0.49\textwidth}
        \centering
        {\fontsize{8}{10}\selectfont
        \setlength{\tabcolsep}{3pt}
        \renewcommand{\arraystretch}{1}
        \begin{tabular}{l r r r r c r}
        \hline
        Target & Model & VAL & UNQ & NOV & $\hat{T}_g$ & MAE \\
        \hline
        \multirow{2}{*}{250}
          & 1M   & 0.9619 & 0.9754 & 0.9981 & $277.5 \pm 33.6$ & 34.8 \\
          & 100M & 0.9904 & 0.9517 & 0.8177 & $271.2 \pm 32.2$ & 30.3 \\
        \hline
        \multirow{2}{*}{350}
          & 1M   & 0.9721 & 0.9894 & 0.9983 & $355.9 \pm 40.3$ & 31.6 \\
          & 100M & 0.9931 & 0.9913 & 0.7445 & $350.5 \pm 37.9$ & 29.7 \\
        \hline
        \multirow{2}{*}{400}
          & 1M   & 0.9739 & 0.9915 & 0.9982 & $401.9 \pm 46.0$ & 36.8 \\
          & 100M & 0.9931 & 0.9923 & 0.6961 & $399.4 \pm 44.5$ & 35.6 \\
        \hline
        \multirow{2}{*}{450}
          & 1M   & 0.9771 & 0.9939 & 0.9980 & $445.1 \pm 46.2$ & 37.4 \\
          & 100M & 0.9937 & 0.9947 & 0.6225 & $444.5 \pm 45.9$ & 37.0 \\
        \hline
        \multirow{2}{*}{500}
          & 1M   & 0.9810 & 0.9950 & 0.9979 & $486.6 \pm 43.3$ & 35.6 \\
          & 100M & 0.9949 & 0.9954 & 0.5707 & $484.9 \pm 42.8$ & 35.9 \\
        \hline
        \multirow{2}{*}{600}
          & 1M   & 0.9774 & 0.9922 & 0.9942 & $553.0 \pm 36.3$ & 50.7 \\
          & 100M & 0.9939 & 0.9937 & 0.4883 & $556.4 \pm 36.5$ & 48.4 \\
        \hline
        \end{tabular}
        }
        \captionof{table}{$T_g$ (K) sweep generation quality: 1M vs.\ 100M training.}
        \label{tab:size_comparison}
    \end{minipage}
\end{figure}

Figures~\ref{fig:oneM_vs_hundredM_pred}, \ref{fig:oneM_vs_hundredM_gen}, and \ref{fig:oneM_vs_hundredM_mae}
show distributions, generation statistics, and MAE figures on different $T_g$ sweeps for the 1M and 100M models.
Full details are provided in Table~\ref{tab:size_comparison}. The 100M model yields slightly lower MAEs and
predicted means closer to the desired $T_g$ values. Validity and uniqueness remain high for both models but are
best for the 100M models. Novelty drops off for the larger training corpus, which is expected as there are more
polymers within the training set that generated structures may be compared to.

\clearpage

\subsection{Model Size}\label{append_modelsize}

\begin{figure}[!htbp]
    \centering
    \includegraphics[width=\linewidth]{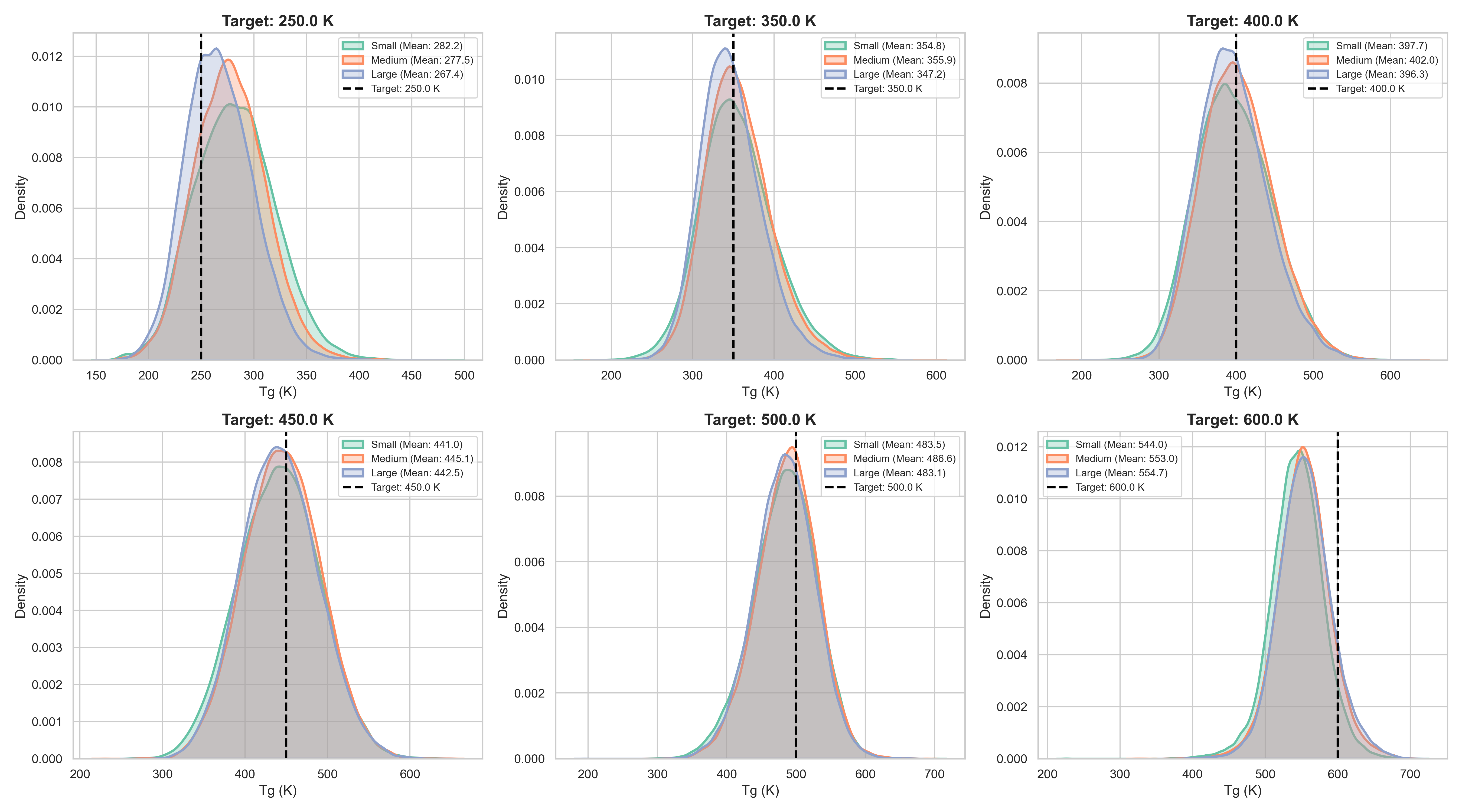}
    \caption{Predicted $T_g$ distributions for Small, Medium, and Large models
    trained on the 1M corpus, across $T_g$ sweep targets.}
    \label{fig:modelsize_1m_pred}
\end{figure}

\begin{figure}[!htbp]
    \centering
    \includegraphics[width=\linewidth]{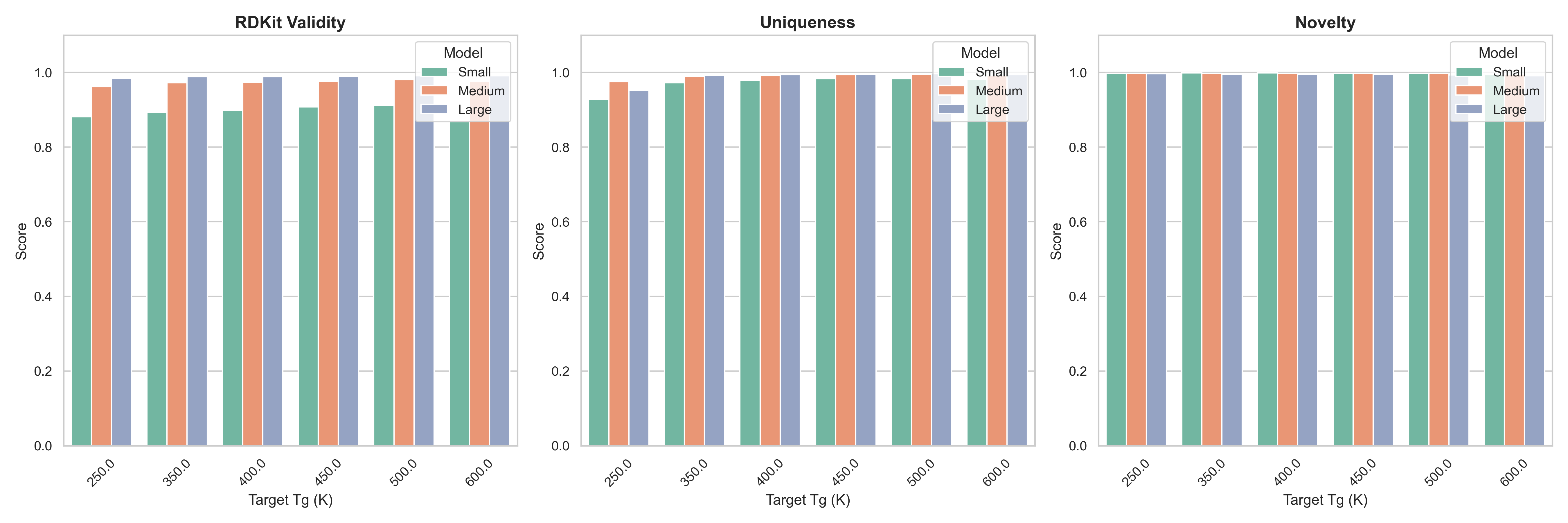}
    \caption{Generation statistics (validity, uniqueness, novelty) for Small,
    Medium, and Large models trained on the 1M corpus, across $T_g$ sweep
    targets.}
    \label{fig:modelsize_1m_gen}
\end{figure}

\begin{figure}[!htbp]
    \centering
    \begin{minipage}{0.49\textwidth}
        \centering
        \includegraphics[width=\linewidth]{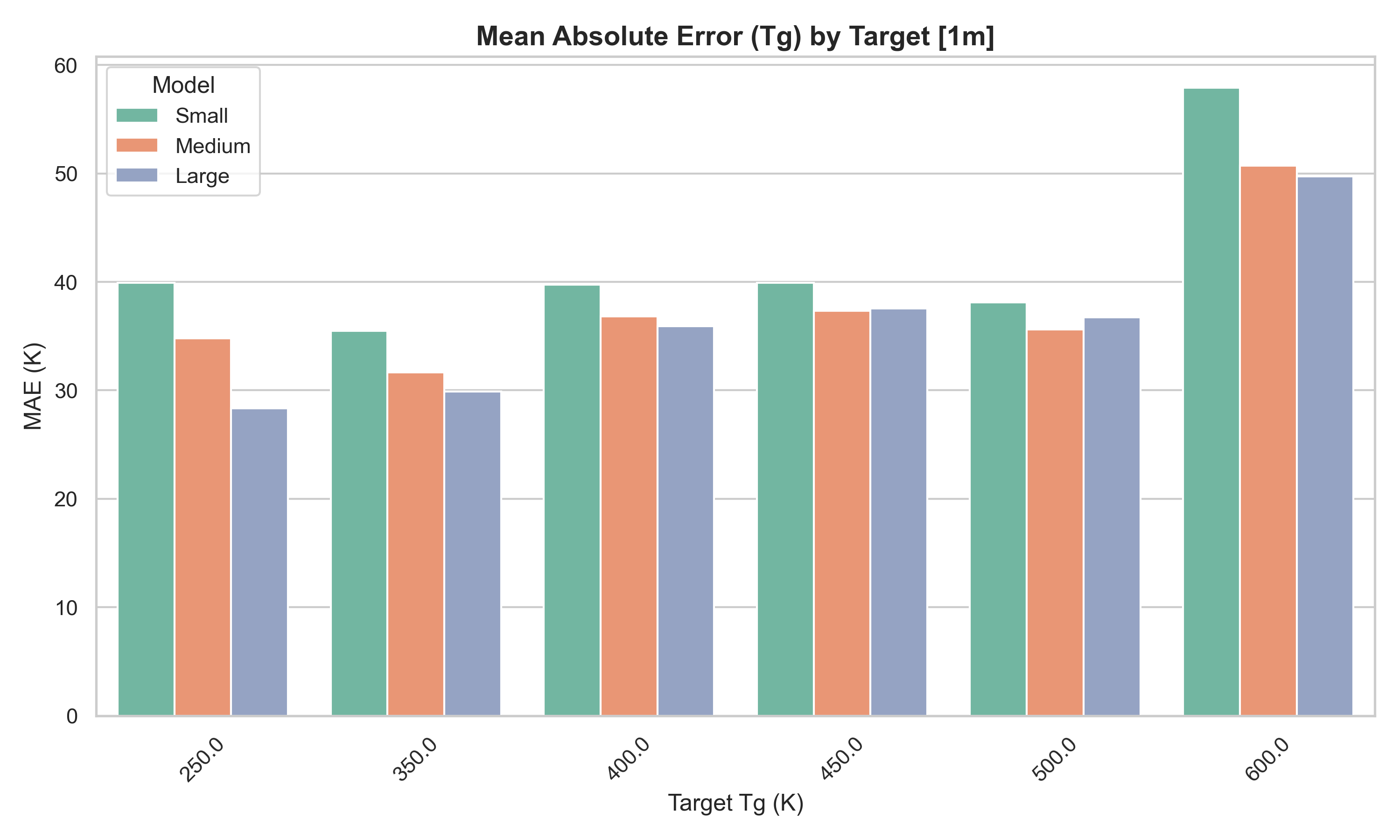}
        \caption{MAE of predicted $T_g$ for Small, Medium, and Large models trained
        on the 1M corpus, across $T_g$ sweep targets.}
        \label{fig:modelsize_1m_mae}
    \end{minipage}\hfill
    \begin{minipage}{0.49\textwidth}
        \centering
        {\fontsize{9}{11}\selectfont
        \setlength{\tabcolsep}{3pt}
        \renewcommand{\arraystretch}{1.05}
        \begin{tabular}{l l r r r c r}
        \hline
        Target & Model & VAL & UNQ & NOV & $\hat{T}_g$ & MAE \\
        \hline
        \multirow{3}{*}{250}
          & Small  & 0.8812 & 0.9291 & 0.9980 & $282.2 \pm 37.8$ & 39.9 \\
          & Medium & 0.9619 & 0.9754 & 0.9981 & $277.5 \pm 33.6$ & 34.8 \\
          & Large  & 0.9851 & 0.9530 & 0.9963 & $267.4 \pm 31.8$ & 28.4 \\
        \hline
        \multirow{3}{*}{350}
          & Small  & 0.8936 & 0.9720 & 0.9984 & $354.8 \pm 45.2$ & 35.5 \\
          & Medium & 0.9721 & 0.9894 & 0.9983 & $355.9 \pm 40.3$ & 31.6 \\
          & Large  & 0.9889 & 0.9926 & 0.9959 & $347.2 \pm 37.7$ & 29.9 \\
        \hline
        \multirow{3}{*}{400}
          & Small  & 0.8987 & 0.9787 & 0.9984 & $397.7 \pm 49.2$ & 39.8 \\
          & Medium & 0.9739 & 0.9915 & 0.9982 & $402.0 \pm 46.0$ & 36.8 \\
          & Large  & 0.9887 & 0.9940 & 0.9957 & $396.3 \pm 44.7$ & 35.9 \\
        \hline
        \multirow{3}{*}{450}
          & Small  & 0.9073 & 0.9830 & 0.9983 & $441.0 \pm 48.9$ & 39.9 \\
          & Medium & 0.9771 & 0.9939 & 0.9980 & $445.1 \pm 46.2$ & 37.4 \\
          & Large  & 0.9901 & 0.9954 & 0.9946 & $442.5 \pm 46.0$ & 37.6 \\
        \hline
        \multirow{3}{*}{500}
          & Small  & 0.9114 & 0.9835 & 0.9978 & $483.5 \pm 45.6$ & 38.1 \\
          & Medium & 0.9810 & 0.9950 & 0.9979 & $486.6 \pm 43.3$ & 35.6 \\
          & Large  & 0.9911 & 0.9964 & 0.9936 & $483.1 \pm 43.2$ & 36.7 \\
        \hline
        \multirow{3}{*}{600}
          & Small  & 0.8682 & 0.9813 & 0.9938 & $544.0 \pm 35.8$ & 57.9 \\
          & Medium & 0.9774 & 0.9922 & 0.9942 & $553.0 \pm 36.3$ & 50.7 \\
          & Large  & 0.9907 & 0.9939 & 0.9907 & $554.7 \pm 36.6$ & 49.7 \\
        \hline
        \end{tabular}
        } % end 9pt group
        \captionof{table}{$T_g$ (K) sweep generation quality by model size and target condition (1M training).}
        \label{tab:size_tg_sweep}
    \end{minipage}
\end{figure}

\textbf{1M Training: Small, Medium, and Large} Figures~\ref{fig:modelsize_1m_pred}, ~\ref{fig:modelsize_1m_gen}, and ~\ref{fig:modelsize_1m_mae} show predicted $T_g$ distributions, generation statistics, and accuracy for Small, Medium, and Large models trained on the 1M corpus and conditioned on $T_g$ only. Quantitative results are summarized in Table~\ref{tab:size_tg_sweep}. As model size increases, validity and uniqueness increase while MAE decreases.

\clearpage

\begin{figure}[!htbp]
    \centering
    \includegraphics[width=\linewidth]{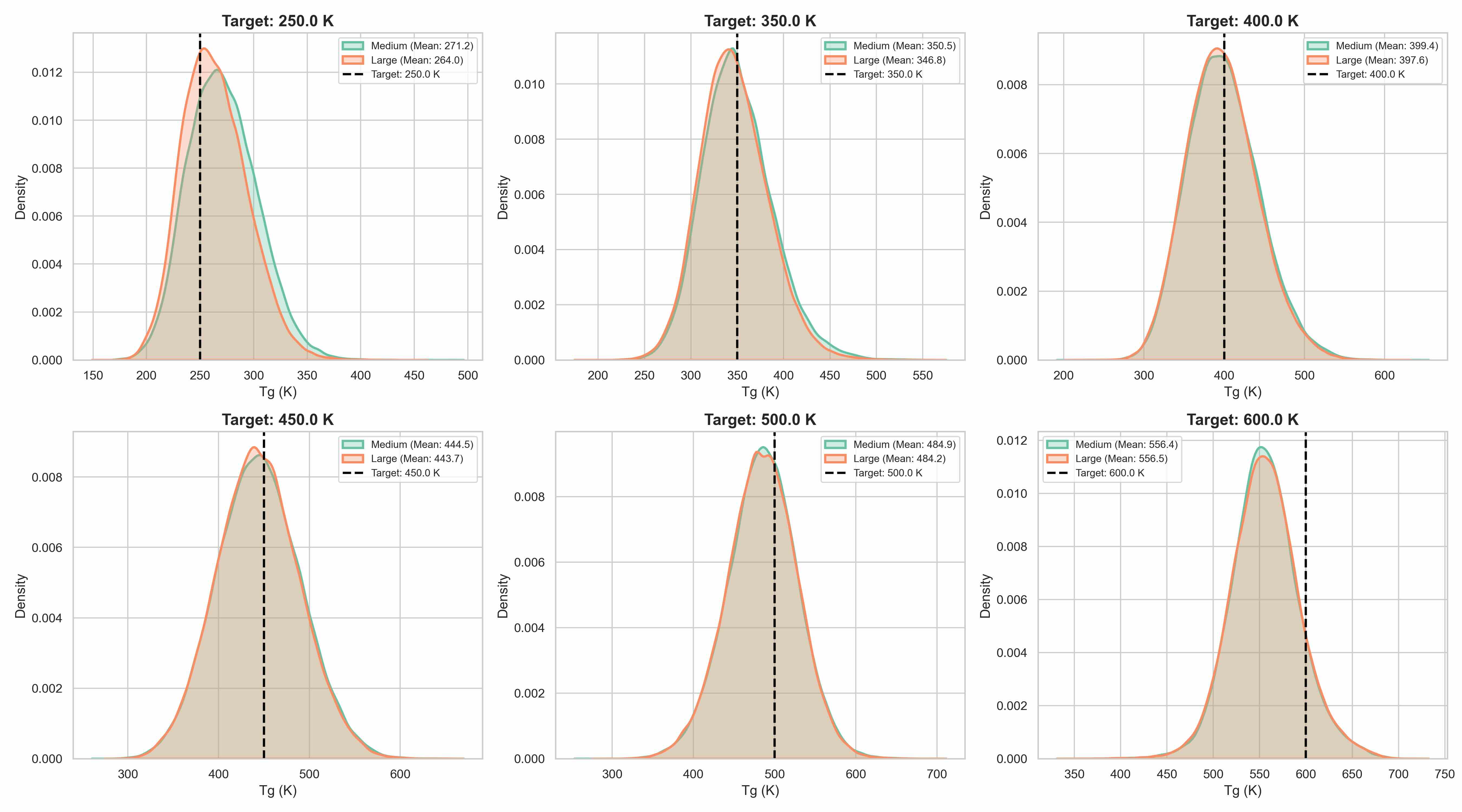}
    \caption{Predicted $T_g$ distributions for Medium and Large models trained
    on the 100M corpus, across $T_g$ sweep targets.}
    \label{fig:modelsize_100m_pred}
\end{figure}

\begin{figure}[H]
    \centering
    \includegraphics[width=.8\linewidth]{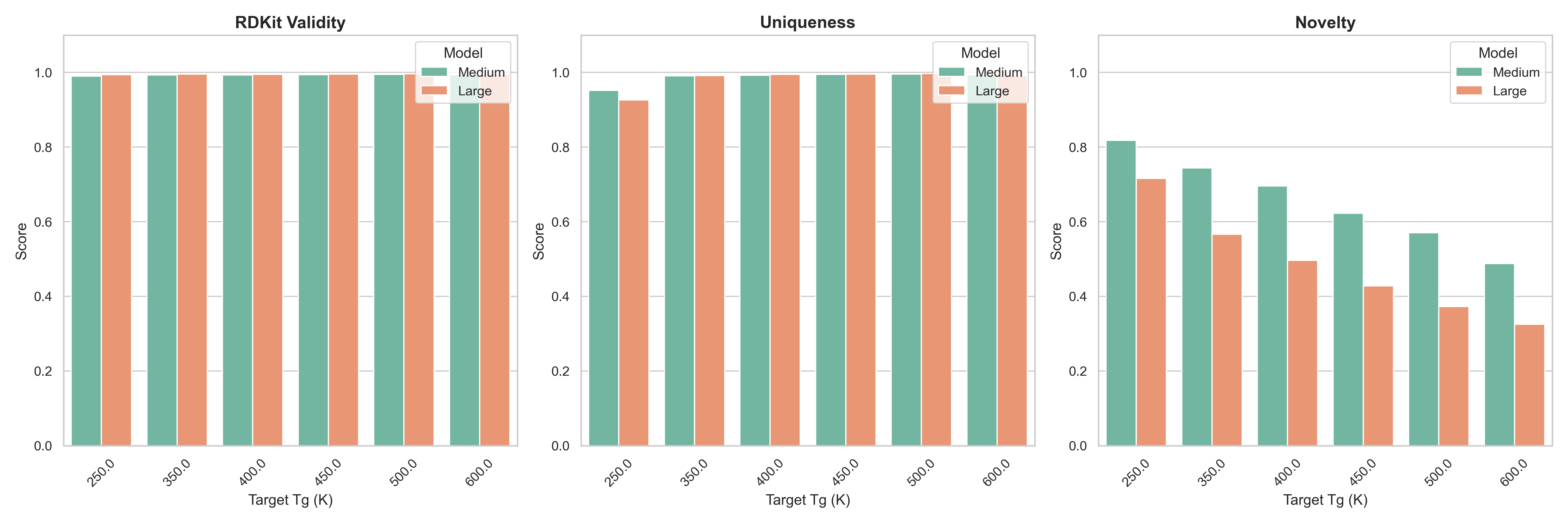}
    \caption{Generation statistics (validity, uniqueness, and novelty) for
    Medium and Large models trained on the 100M corpus, across $T_g$ sweep
    targets.}
    \label{fig:modelsize_100m_gen}
\end{figure}

\begin{figure}[H]
    \centering
    \begin{minipage}{0.49\textwidth}
        \centering
        \includegraphics[width=\linewidth]{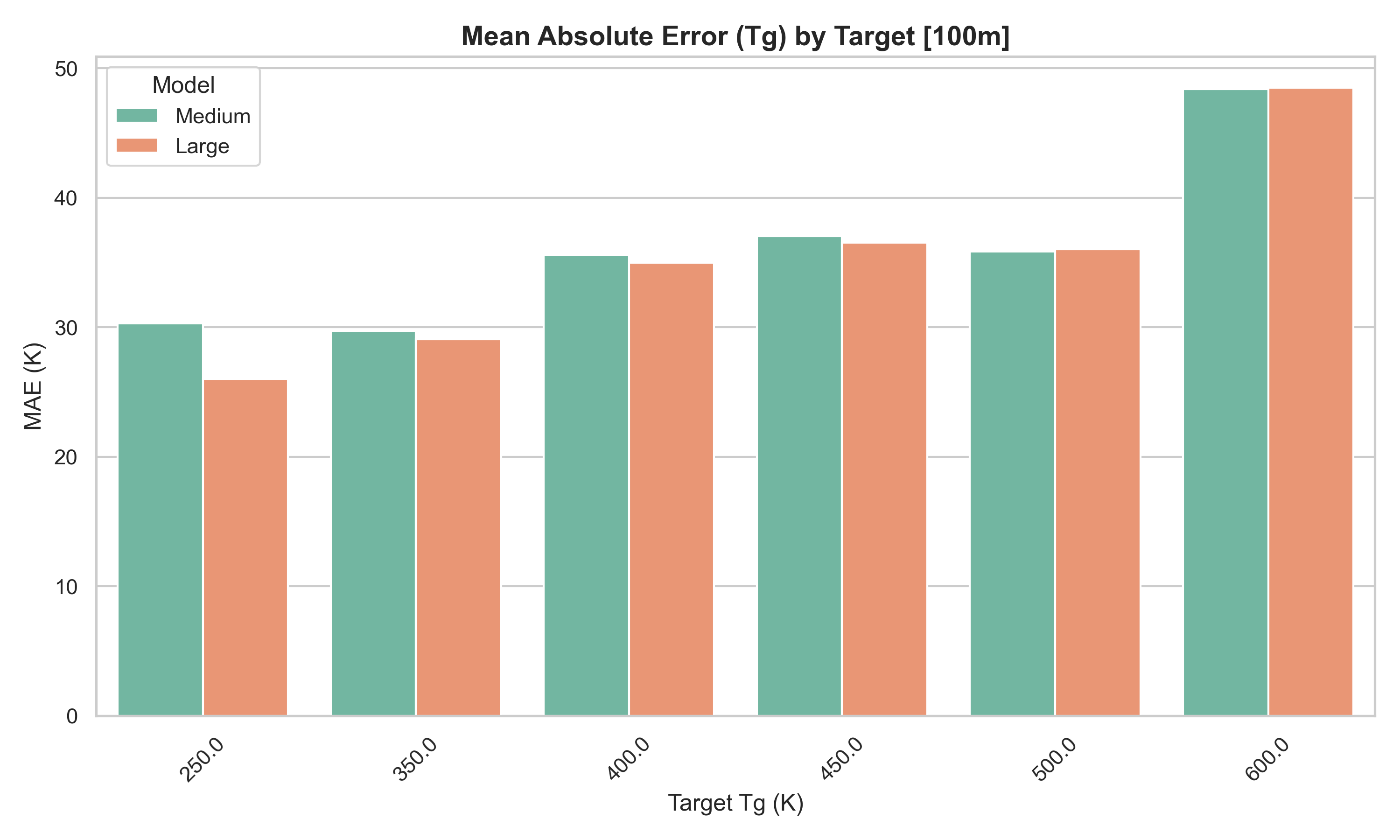}
        \caption{Mean absolute error (MAE) of predicted $T_g$ for Medium and Large
        models trained on the 100M corpus, across $T_g$ sweep targets.}
        \label{fig:modelsize_100m_mae}
    \end{minipage}\hfill
    \begin{minipage}{0.49\textwidth}
        \centering
        {\fontsize{9}{11}\selectfont
        \setlength{\tabcolsep}{3pt}
        \renewcommand{\arraystretch}{1.05}
        \begin{tabular}{l l r r r c r}
        \hline
        Target & Model & VAL & UNQ & NOV & $\hat{T}_g$  & MAE  \\
        \hline
        \multirow{2}{*}{250 K}
          & Medium & 0.9904 & 0.9517 & 0.8177 & $271.2 \pm 32.2$ & 30.3 \\
          & Large  & 0.9938 & 0.9261 & 0.7164 & $264.0 \pm 30.3$ & 26.0 \\
        \hline
        \multirow{2}{*}{350 K}
          & Medium & 0.9931 & 0.9913 & 0.7445 & $350.5 \pm 37.9$ & 29.7 \\
          & Large  & 0.9956 & 0.9914 & 0.5665 & $346.8 \pm 36.4$ & 29.1 \\
        \hline
        \multirow{2}{*}{400 K}
          & Medium & 0.9931 & 0.9923 & 0.6961 & $399.4 \pm 44.5$ & 35.6 \\
          & Large  & 0.9952 & 0.9951 & 0.4969 & $397.6 \pm 43.6$ & 35.0 \\
        \hline
        \multirow{2}{*}{450 K}
          & Medium & 0.9937 & 0.9947 & 0.6225 & $444.5 \pm 45.9$ & 37.0 \\
          & Large  & 0.9956 & 0.9959 & 0.4279 & $443.7 \pm 45.1$ & 36.5 \\
        \hline
        \multirow{2}{*}{500 K}
          & Medium & 0.9949 & 0.9954 & 0.5707 & $484.9 \pm 42.8$ & 35.9 \\
          & Large  & 0.9965 & 0.9974 & 0.3728 & $484.2 \pm 42.7$ & 36.0 \\
        \hline
        \multirow{2}{*}{600 K}
          & Medium & 0.9939 & 0.9937 & 0.4883 & $556.4 \pm 36.5$ & 48.4 \\
          & Large  & 0.9956 & 0.9923 & 0.3253 & $556.5 \pm 37.1$ & 48.5 \\
        \hline
        \end{tabular}
        } % end 9pt group
        \captionof{table}{$T_g$ (K) sweep generation quality for Medium and Large models (100M training).}
        \label{tab:100M_tg_sweep}
    \end{minipage}
\end{figure}

\textbf{100M Training: Medium, and Large}
Figure~\ref{fig:modelsize_100m_pred} shows the predicted $T_g$ distributions for Medium and Large models trained on the 100M corpus across the same sweep targets. Figure~\ref{fig:modelsize_100m_mae} shows the corresponding MAE across sweep targets, and Figure~\ref{fig:modelsize_100m_gen} summarizes generation statistics (validity, uniqueness, and novelty) for each condition. Table~\ref{tab:100M_tg_sweep} summarizes generation quality and
prediction accuracy for each sweep condition and model size.

\clearpage

\subsection{Generation Sizes}\label{append_gensize}

\begin{figure}[!htbp]
    \centering
    \includegraphics[width=\linewidth]{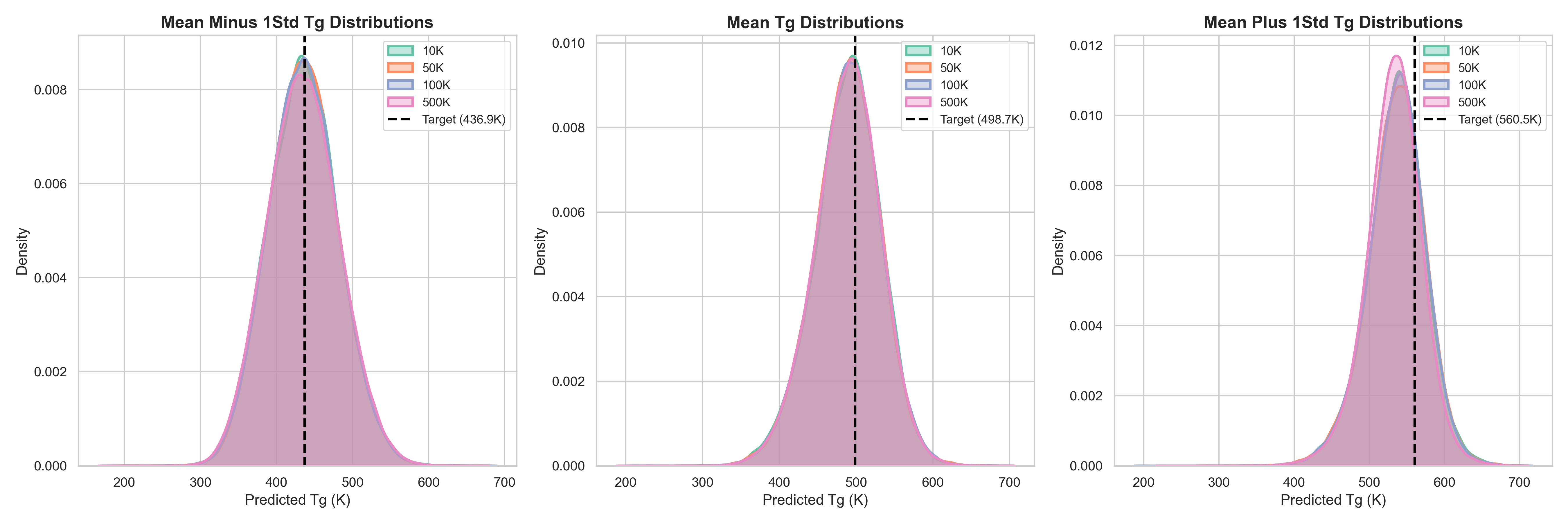}
    \caption{Predicted Tg distributions for generation sizes ranging from 10K to 500K. The model reliably centers the distributions around the targets across all scales.}
    \label{fig:gensize_preddist}
\end{figure}

\begin{figure}[H]
    \centering
    % Left: figure
    \begin{minipage}{0.49\textwidth}
        \centering
        \includegraphics[width=\linewidth]{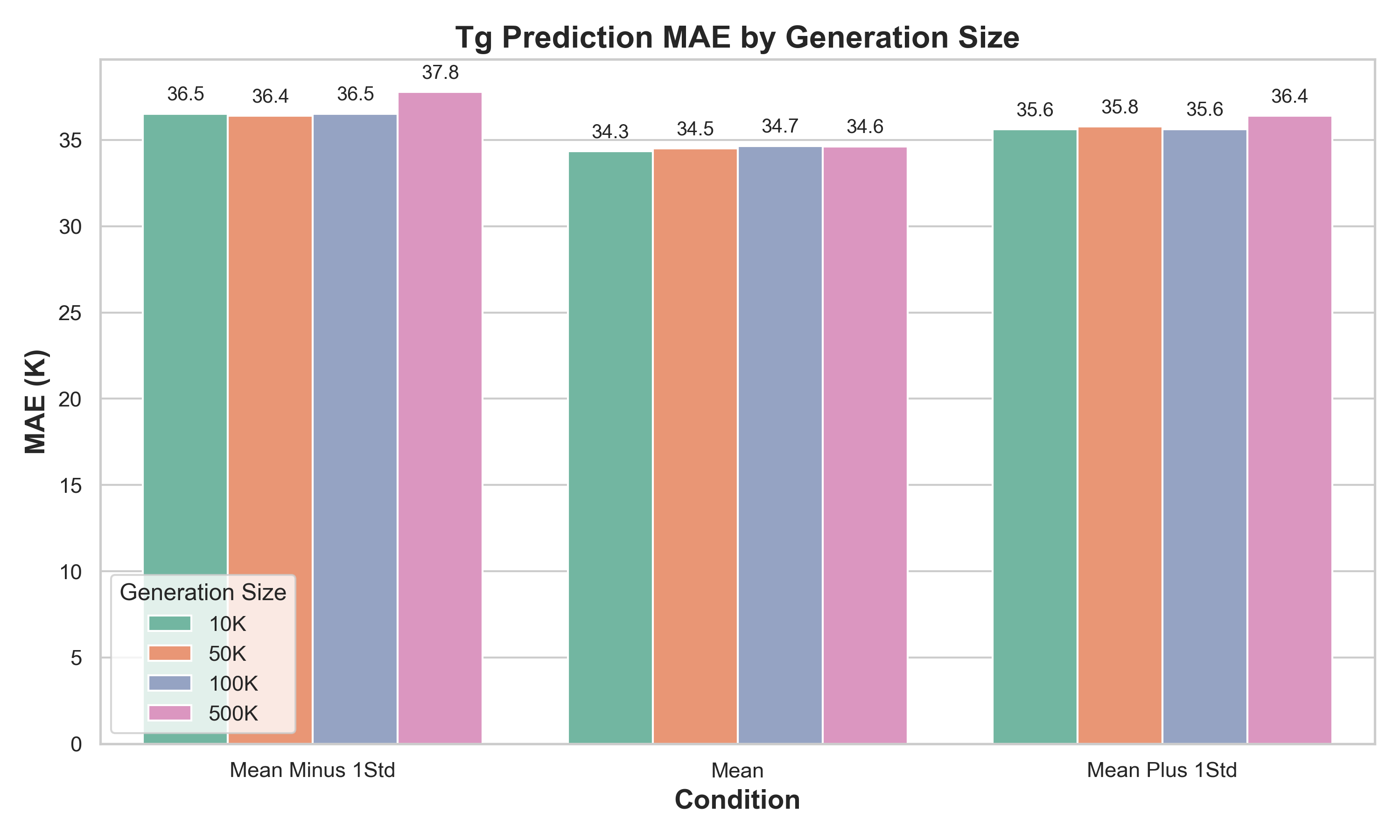}
        \caption{Mean Absolute Error (MAE) of predicted $T_g$ values across different generation sizes. Prediction accuracy remains consistent irrespective of the number of samples generated.}
        \label{fig:gensize_predstats}
    \end{minipage}\hfill
    \begin{minipage}{0.49\textwidth}
        \centering
        {\fontsize{9}{11}\selectfont
        \setlength{\tabcolsep}{3pt}
        \renewcommand{\arraystretch}{1.05}
        \begin{tabular}{l r r r r r r r}
        \hline
        & & \multicolumn{3}{c}{Generation metrics} & \multicolumn{3}{c}{Predicted $T_g$ (K)} \\
        \cline{3-5}\cline{6-8}
        Condition & N & VAL & UNQ & NOV & Mean & Std & MAE \\
        \hline
        \multirow{4}{*}{\shortstack[l]{$\mu - \sigma$ \\ (436.93 K)}}
          & 10K  & 0.9798 & 0.9841 & 0.9977 & 435.6 & 45.6 & 36.5 \\
          & 50K  & 0.9791 & 0.9820 & 0.9972 & 435.6 & 45.5 & 36.4 \\
          & 100K & 0.9794 & 0.9804 & 0.9969 & 435.4 & 45.6 & 36.5 \\
          & 500K & 0.9765 & 0.9905 & 0.9983 & 435.0 & 47.0 & 37.8 \\
        \hline
        \multirow{4}{*}{\shortstack[l]{$\mu$ \\ (498.71 K)}}
          & 10K  & 0.9849 & 0.9871 & 0.9962 & 488.6 & 42.7 & 34.3 \\
          & 50K  & 0.9829 & 0.9864 & 0.9965 & 488.1 & 42.7 & 34.5 \\
          & 100K & 0.9841 & 0.9848 & 0.9960 & 488.3 & 42.9 & 34.7 \\
          & 500K & 0.9844 & 0.9765 & 0.9965 & 488.3 & 42.8 & 34.6 \\
        \hline
        \multirow{4}{*}{\shortstack[l]{$\mu + \sigma$ \\ (560.49 K)}}
          & 10K  & 0.9849 & 0.9915 & 0.9926 & 535.9 & 38.4 & 35.6 \\
          & 50K  & 0.9864 & 0.9883 & 0.9937 & 536.0 & 38.8 & 35.8 \\
          & 100K & 0.9855 & 0.9849 & 0.9935 & 535.9 & 38.5 & 35.6 \\
          & 500K & 0.9828 & 0.9908 & 0.9964 & 532.8 & 36.8 & 36.4 \\
        \hline
        \end{tabular}
        } % end 9pt group
        \captionof{table}{$T_g$ sweep generation quality and prediction accuracy across generation
        sizes (1M model, conditioned on $T_g$ only).}
        \label{tab:supp_gensizes}
    \end{minipage}
\end{figure}

Figure~\ref{fig:gensize_preddist} shows predicted property distributions across generation sizes.
Figure~\ref{fig:gensize_predstats} shows prediction statistics across generation sizes.
Table~\ref{tab:supp_gensizes} summarizes statistics for all three sweeps across 4 generation sizes.

\section{Runtimes}\label{append_runtimes}

\subsection{Training}
Table \ref{tab:generator_training} summarizes the training times for the PolymerGPT generator across training corpus size (1M or 100M), model size (small, medium, large), and conditioning setup. Table \ref{tab:predictor_training} summarizes the training times for the TransPolymer property prediction model. Finetuning for the $T_g$ model took 7 hours and 40 minutes using 2 Nvidia A100 GPUs.

\begin{table}[h!]
\centering
\begin{tabular}{lccc}
\hline
\textbf{Model} & \textbf{GPU } & \textbf{Num GPUs} & \textbf{Training Time} \\
\hline
1M small ($T_{g}$ only) &  A100 & 4 & 0:21 \\
1M medium (uncond.) &  L4 & 8 & 14:30 \\
1M medium ($T_{g}$ only) &  A100 & 8 & 0:27 \\
1M medium (2 props) &  A100 & 8 & 0:27 \\
1M medium (3 props) &  A100 & 8 & 0:27 \\
1M medium (37 props) &  L4 & 8 & 2:47 \\
1M medium (scaff.) &  L4 & 4 &  4:01\\
1M medium (scaff. + $T_{g}$)  &  L4 & 4 &  4:02\\
1M large ($T_{g}$ only) &  T4 & 4 & 0:41 \\
100M medium ($T_{g}$ only) &  L4 & 8 & 1:20:51 \\
100M large ($T_{g}$ only) &  A100 & 4 & 2:03:52 \\
\hline
\end{tabular}
\caption{Training times for PolymerGPT generator models. Times are reported as days:hours:minutes where applicable, or hours:minutes otherwise.}
\label{tab:generator_training}
\end{table}

\begin{table}[h!]
\centering
\begin{tabular}{lccc}
\hline
\textbf{Model} & \textbf{GPU Name} & \textbf{Num GPUs} & \textbf{Training Time} \\
\hline
$T_{g}$ model &  A100 & 1 & 2:00:50 \\
$e_{g}^{c}$ model &  A100 & 1 & 0:43:18 \\
$e_{g}^{b}$ model &  A100 & 1 & 0:41:44 \\
$X_{c}$ model &  A100 & 1 & 0:30:41 \\
$e_{\mathrm{ea}}$ model &  A100 & 1 & 0:27:10 \\
\hline
\end{tabular}
\caption{Training times for TransPolymer property predictor models fine-tuned on each of the five target properties.}
\label{tab:predictor_training}
\end{table}

\subsection{Inference}
Across multiple different runs using the 1M 37‑conditional model, we found the average generation time for 50,000 structures to be 10 minutes and 25 seconds, when using a T4 gpu. The average generated sequence length of 105.76 characters gives us an effective token generation rate of approximately 8,090 characters per second. In Tables \ref{tab:generation_times} and \ref{tab:prediction_times}, we display timing and configurations for different generation and prediction runs. It it important to note that these times include full job submission timing including gpu provisioning, code downloading, and other orchestration overheads. As such and because different runs used different generation sizes and gpu counts, reported times should be interpreted as end‑to‑end pipeline latencies rather than pure model throughput, and direct  comparisons should be made with care.

\begin{table}[h!]
\centering
\begin{tabular}{lccccc}
\hline
\textbf{Model} & \textbf{GPU} & \textbf{Num GPUs} & \textbf{Time} & \textbf{Num Structures} & \textbf{Structs/s} \\
\hline
1M small ($T_{g}$ only) & A100 & 1 & 0:33 & 6$\times$50K & 151.5 \\
1M medium ($T_{g}$ only) & T4 & 1 & 1:16 & 6$\times$50K & 65.8 \\
1M large ($T_{g}$ only) & A100 & 1 & 0:59 & 6$\times$50K & 84.7 \\
100M medium ($T_{g}$ only) & A100 & 1 & 0:57 & 6$\times$50K & 87.7 \\
100M large ($T_{g}$ only) & A100 & 1 & 1:23 & 6$\times$50K & 60.2 \\
\hline
\end{tabular}
\caption{Generation times and throughput (structures per second) for PolymerGPT. Times are reported as hours:minutes.}
\label{tab:generation_times}
\end{table}

\begin{table}[h!]
\centering
\begin{tabular}{lccccc}
\hline
\textbf{Model} & \textbf{GPU} & \textbf{Num GPUs} & \textbf{Time} & \textbf{Num Structures} & \textbf{Structs/s} \\
\hline
Property Prediction ($T_{g}$ only) &  A100 & 1 & 0:03 & 50K & 16{,}666.7 \\
Property Prediction (5 properties) &  A100 & 1 & 0:07 & 50K & 7{,}142.9 \\
\hline
\end{tabular}
\caption{Prediction times and throughput for TransPolymer. Times are reported as hours:minutes}
\label{tab:prediction_times}
\end{table}

\section{PolymerGPT Training and Optimization Details}

Training uses teacher-forced next-token prediction over SMILES positions
only,

{\fontsize{9}{\baselineskip}\selectfont
\begin{equation}
  \mathcal{L} = -\frac{1}{N} \sum_{i=1}^{N} \log P_\theta\!\left(x_i \mid x_{<i}, \mathbf{p}\right),
  \label{eq:loss}
\end{equation}}
optimized with AdamW ($\beta_1=0.9$, $\beta_2=0.95$, weight decay $0.1$
applied only to \code{nn.Linear} weight matrices, following standard GPT
practice), gradient clipping at norm 1.0, and mixed-precision (float16
forward/loss, float32 parameters and optimizer state). Learning rate follows
linear warmup (10\% of tokens) then cosine decay to 10\% of peak, with peak
values scaled from a $B{=}512$ base via $\mathrm{lr}_{\mathrm{peak}} = 1.2\times10^{-3}\sqrt{B/512}$ (Table~\ref{tab:lr}). The loss is the mean cross-entropy over all
SMILES token positions, with the property prefix excluded (Eq. \ref{eq:loss}). Although the property prefix's own logit row is excluded from the loss, its hidden state participates in every subsequent token's attention computation, so $\mathrm{prop\_nn}$ receives gradient signal from all $L$ downstream predictions rather than any direct reconstruction target.

\begin{table}[h]
\centering
\fontsize{9}{12}\selectfont
\setlength{\tabcolsep}{4pt}
\renewcommand{\arraystretch}{1.05}
\begin{tabular}{l c c c}
\hline
Variant & Hardware & Batch per rank & Peak LR \\
\hline
Small &  4$\times$A100 40GB  & 512 & 6$\times10^{-4}$ \\
Medium & 8$\times$L4 24GB & 256  & $8.49\times10^{-4}$ \\
Large  & 4$\times$A100 40GB & 512 & $1.20\times10^{-3}$ \\
\hline
\end{tabular}
\caption{Learning-rate configuration by model variant.}
\label{tab:lr}
\end{table}

Three model scales are evaluated (Table~\ref{tab:variants}). 

\begin{table}[h]
\centering
\fontsize{9}{12}\selectfont
\setlength{\tabcolsep}{4pt}
\renewcommand{\arraystretch}{1.05}
\begin{tabular}{l c c c c c}
\hline
Variant & $L$ & $H$ & $d$ & MLP dim & Params \\
\hline
Small  & 4  & 4 & 128 & 512  & $\sim$1M  \\
Medium & 8  & 8 & 256 & 1024 & $\sim$7M  \\
Large  & 12 & 8 & 512 & 2048 & $\sim$38M \\
\hline
\end{tabular}
\caption{Three model variants and their detailed settings}
\label{tab:variants}
\end{table}

Models are trained with PyTorch DDP (NCCL backend) across the GPU
configurations in Table~\ref{tab:lr}, with file-level data sharding across
ranks and a global-mean reduction for validation loss. The large model
exceeds A100 40GB memory at useful batch sizes; we apply block-level
gradient checkpointing
% \cite{chen2016training} {\color{red}(fix ref)
, trading a
$\sim$33\% increase in per-step compute for an $\sim$5$\times$ reduction in
peak activation memory (37GB $\to$ 5–7GB), not required for the medium
variant.

{\noindent \bf Pre-Processing data}
The 100M PolyOne dataset uses 500K polymer strings in each parquet file. We utilized the exisitng canonicalization library and created a 250K polymer strings parquet file for convenience. The total number of resulting parquet files are 396, as we lost few strings due to the failure of canonicalization. The subset 1 million psmiles strings are taken from the first four canoncialized parquet files and an additional file for validation. For easy differentiation we name this subset as PolyOne1M while the full data is referred as PolyOne100M.

{\noindent \bf Streaming vs. In-Memory data}
To avoid out-of-memory error on RAM by the training data, we employ streaming of data rather than in-memory data. Although PolyOne1M can be loaded directly into RAM, for consistency and scalability we have employed streaming of these canonicalized psmile strings as default. 

{\noindent \bf Multi-GPU}
The computational resources required to train a model on huge data are significant. Utilizing the Gemini Enterprise Agent Platform's ability to provision multiple GPUs for each node, we used Distributed Data Parallel from the PyTorch to replicate the model on each GPU. For the backward pass, we aggregate all the loss values and therfore the model weights are updated from all the GPUs. Specifically, for training PolyOne100M data we used 8 GPUs of NVIDIA L4 or A100 depending on the successful provisioning.

{\noindent \bf Distributed Training}
Memory pressure at this scale came from two distinct sources — training-time
activation memory and generation-time recomputation — and we addressed them
with two separate mechanisms, since the training and generation code paths
use independent model implementations.

\paragraph{Flash attention (training).} 
\label{sec: flash} The training model's self-attention layer is a custom
multi-head implementation — separate query/key/value projections and
explicit head reshaping, not \texttt{torch.nn.MultiheadAttention} — but the
core attention computation is delegated to
\texttt{torch.nn.functional.scaled\_dot\_product\_attention} with
\texttt{is\_causal=True}, which lets PyTorch dispatch that call to a fused
Flash-Attention kernel instead of explicitly materializing the $O(N^2)$
attention matrix. This keeps attention memory at $O(N)$ in sequence length,
which is what makes the large variant's peak activation memory tractable
enough for gradient checkpointing to close the
remaining gap on a 40GB A100.

\paragraph{KV caching (generation).} The generation model is a separate
implementation that does not use fused attention; instead it caches
per-layer key/value tensors and, after an initial
full-context forward pass over the conditioning prefix and prompt, feeds
only the newly generated token at each subsequent step, concatenating its
key/value onto the cache rather than recomputing attention over the full
prefix. This reduces per-step generation cost from $O(T)$ to $O(1)$ and
total sampling cost from $O(T^2)$ to $O(T)$ in sequence length $T$ — a
compute rather than a training-memory optimization, but one that made
iterating on sampling (property-conditioned generation runs used throughout
evaluation) practical.

{\noindent \bf Memory Management at Scale} \label{sec:mem}
Even with flash attention holding attention memory at $O(N)$, the large variant's per-layer activations
still exceeded a 40GB A100 at useful batch sizes: an initial run at batch
512 peaked at $\sim$37GB, leaving no margin for the property-conditioning
prefix, optimizer state, or NCCL communication buffers, and OOM'd in
practice. We applied block-level gradient checkpointing — discarding intermediate block activations on the
forward pass and recomputing them on demand during backward, rather than
retaining all of them for the full backward pass — which brought peak
activation memory down to $\sim$5--7GB at the same batch size, at the cost
of a $\sim$33\% increase in per-step  time from the extra forward
recomputation. This made batch 512 practical on the large variant. Gradient checkpointing is not
required for the medium variant (Table~\ref{tab:variants}), which fits
comfortably within 24GB L4 memory at its trained batch size without it.

{\noindent \bf Sampling and Temperature}
\label{sec:sampling}
Autoregressive sampling from $P_\theta(\mathbf{x}\mid\mathbf{p})$ scales
logits by a temperature $T$ before the softmax at each decoding step,
$P(x_i) \propto \exp(\mathrm{logit}_i / T)$, optionally followed by
top-$k$ truncation to the $k$ highest-probability tokens before renormalizing.
$T{\to}0$ concentrates mass on the model's most confident continuation
(approaching greedy decoding), favoring syntactically valid, canonicalizable
pSMILES at the cost of diversity, since the model repeatedly favors common
training-set motifs; $T{>}1$ flattens the distribution and increases
structural diversity, at the cost of a higher rate of invalid sequences
(unbalanced ring-closure or bracket tokens) from tail tokens that are only
weakly supported at a given decoding step. Top-$k$ truncation mitigates this
by bounding the sampling pool independently of $T$, trading off diversity
for a hard cap on how unlikely a sampled token can be.

We use two different operating points for these knobs, matched to two
different purposes. The per-epoch in-training sampling check
samples at $T{=}0.8$ with $k{=}10$ — a conservative setting that keeps the
diagnostic itself validity-biased so it reflects model progress rather than
sampling noise. Final generation for evaluation, by contrast, samples at
$T{=}1.0$ with no top-$k$ truncation across all conditional and
unconditional generation scripts, i.e.\ directly from the model's learned
distribution over the full vocabulary at each step, favoring diversity and
faithfulness to $P_\theta$ over a validity floor; any invalid sequences are
filtered post-hoc rather than suppressed at sampling time.
\clearpage

\section{TransPolymer Fine-tuning}\label{sec:appendix_transpolymer}

\textbf{Datasets}
Table~\ref{tab:Pred_finetine_sources} refers to the sources and data sizes for the property prediction model. Each dataset is split and trained using a 5-fold cross-validation split. Each dataset contains an augmented size created through SMILES enumeration. This is done by generating multiple non-canonical variations for each original polymer sequence by systematically rotating the starting atom index, renumbering the atoms, and generating new non-canonical SMILES strings using RDKit. We then remove any duplicates and randomly sample these structurally equivalent but textually different strings to reach the target augmented dataset size. The custom dataset for the $T_g$ dataset was taken from \cite{awguhst2023polymertg}.

\begin{table*}[!htbp]
\centering
\fontsize{10}{12}\selectfont
\setlength{\tabcolsep}{3pt}
\renewcommand{\arraystretch}{1.05}
\begin{tabular}{l l r r r}
\hline
Symbol & Source & \# Data & \# Augmented Train & \# Test \\
\hline
1: $e_{g}^{c}$        & TransPolymer (Egc) & 3,380 &  5,408 &    676 \\
2: $e_{g}^{b}$        & TransPolymer (Egb) &   561 &  6,443 &    113 \\
6: $e_{\mathrm{ea}}$  & TransPolymer (Eea) &   368 &  3,993 &     74 \\
32: $X_{c}$           & TransPolymer (Xc)  &   432 &  8,837 &     87 \\
36: $T_{g}$           & Custom Training    & 7,284 & 17,387 &  1,457 \\
\hline
\end{tabular}
\caption{Dataset information for the property prediction model.}
\label{tab:Pred_finetine_sources}
\end{table*}

\textbf{Hyperparameter Selection} 
Table~\ref{tab:hp_per_property} shows the optimal hyperparameters found for each property predictor. The first four properties use the fine-tuned parameters from the TransPolymer paper, while the custom-trained $T_g$ model was optimized using Ray Tune with the Bayesian Optimization and HyperBand (BOHB) search algorithm. Optimal batch size was found to be 64 across all properties. Additionally, the augmentation limit column shows the maximum amount of augmented pSMILES allowed to be generated per training point. Table~\ref{tab:hp_search_space} shows the respective hyperparameters along with their search type and the spaces searched for.

\begin{table*}[!htbp]
\centering
\fontsize{10}{12}\selectfont
\setlength{\tabcolsep}{4pt}
\renewcommand{\arraystretch}{1.05}
\begin{tabular}{l r r r r r r r r}
\hline
Symbol & lr\_rate & lr\_rate\_reg & weight\_decay & warmup\_ratio & hidden\_dropout & attn\_dropout & drop\_rate & Aug Limit \\
\hline
1: $e_{g}^{c}$       & 1.0e-4  & 1.0e-4  & 0.01    & 0.1     & 0.1     & 0.1     & 0.1     & 2    \\
2: $e_{g}^{b}$       & 5.0e-5  & 5.0e-5  & 0.01    & 0.1     & 0.1     & 0.1     & 0.1     & ---  \\
6: $e_{\mathrm{ea}}$ & 5.0e-5  & 5.0e-5  & 0.01    & 0.1     & 0.1     & 0.1     & 0.1     & ---  \\
32: $X_{c}$          & 5.0e-5  & 5.0e-5  & 0.01    & 0.1     & 0.1     & 0.1     & 0.1     & ---  \\
36: $T_{g}$          & 9.17e-5 & 1.79e-5 & 7.99e-5 & 0.0749  & 0.4887  & 0.3075  & 0.2455  & 3    \\
\hline
\end{tabular}
\caption{Hyperparameters and augmentation limit per property symbol.}
\label{tab:hp_per_property}
\end{table*}

\begin{table*}[!htbp]
\centering
\fontsize{10}{12}\selectfont
\setlength{\tabcolsep}{4pt}
\renewcommand{\arraystretch}{1.05}
\begin{tabular}{l l l}
\hline
Hyperparameter & Search Type   & Range / Values \\
\hline
batch\_size                     & Choice       & [16, 32, 64] \\
lr\_rate                        & Log-Uniform & 1e-5 to 1e-4 \\
lr\_rate\_reg                   & Log-Uniform & 1e-5 to 1e-4 \\
weight\_decay                   & Log-Uniform & 1e-5 to 1e-2 \\
warmup\_ratio                   & Uniform     & 0.05 to 0.1 \\
hidden\_dropout\_prob           & Uniform     & 0.1 to 0.5 \\
attention\_probs\_dropout\_prob & Uniform     & 0.1 to 0.5 \\
drop\_rate                      & Uniform     & 0.1 to 0.5 \\
\hline
\end{tabular}
\caption{Hyperparameter search space used for TransPolymer fine-tuning.}
\label{tab:hp_search_space}
\end{table*}

\pagebreak

\end{document}